\documentclass{article}
\usepackage{titletoc}
\usepackage[nonatbib,final]{neurips_2021}\usepackage[nonatbib]{neurips_2021}
\usepackage[square,numbers]{natbib}
\usepackage[utf8]{inputenc} %
\usepackage[T1]{fontenc}    %
\usepackage{hyperref}       %
\usepackage{url}            %
\usepackage{booktabs}       %
\usepackage{amsfonts}       %
\usepackage{nicefrac}       %
\usepackage{microtype}      %
\usepackage{xcolor}         %
\usepackage{graphicx} 
\usepackage{wrapfig}

\renewcommand*{\thefootnote}{\fnsymbol{footnote}}
\title{Why Do Better Loss Functions\\Lead to Less Transferable Features?}
\author{%
  Simon Kornblith{\normalfont\textsuperscript{1}}\footnotemark[1]\,\,, Ting Chen{\normalfont\textsuperscript{1}}\,, Honglak Lee{\normalfont\textsuperscript{2}}\footnotemark[2]\,\,, Mohammad Norouzi{\normalfont\textsuperscript{1}}\\
  \textsuperscript{1}Google Research, Toronto \textsuperscript{2}University of Michigan
}
\renewcommand*{\thefootnote}{\arabic{footnote}}

\usepackage{amsfonts}       %
\usepackage{amsmath}
\usepackage{bm}
\usepackage{chngcntr}
\usepackage{placeins}
\usepackage{enumitem}
\usepackage{bibunits}

\usepackage[font=small]{caption}
\newcommand\textcite[1]{\citet{#1}}
\newcommand\aftertablecaption{\vspace{-0.0em}\footnotesize}
\def\figref#1{Figure~\ref{#1}}

\def\tabref#1{Table~\ref{#1}}

\def\eqref#1{(\ref{#1})}

\def\1{\bm{1}}

\newenvironment{talign}
 {\align}
 {\endalign}
\newenvironment{talign*}
 {\csname align*\endcsname}
 {\endalign}

\newcommand{\cutsectionup}{}
\newcommand{\cutsectiondown}{}
\newcommand{\cutsubsectionup}{}
\newcommand{\cutsubsectiondown}{}
\begin{document}
\maketitle

\begin{bibunit}[abbrvnat]
\renewcommand\cite[1]{\citep{#1}}
\renewcommand*{\thefootnote}{\fnsymbol{footnote}}
\footnotetext[1]{Correspondence to: \texttt{\href{mailto:skornblith@google.com}{skornblith@google.com}}}
\footnotetext[2]{Work performed while at Google.}
\renewcommand*{\thefootnote}{\arabic{footnote}}
\begin{abstract}
Previous work has proposed many new loss functions and regularizers that improve test accuracy on image classification tasks. However, it is not clear whether these loss functions learn better representations for downstream tasks.
This paper studies how the choice of training objective affects the transferability of the hidden representations of convolutional neural networks trained on ImageNet. We show that many objectives lead to statistically significant improvements in ImageNet accuracy over vanilla softmax cross-entropy, but the resulting fixed feature extractors transfer substantially worse to downstream tasks, and the choice of loss has little effect when networks are fully fine-tuned on the new tasks. Using centered kernel alignment to measure similarity between hidden representations of networks, we find that differences among loss functions are apparent only in the last few layers of the network. We delve deeper into representations of the penultimate layer, finding that different objectives and hyperparameter combinations lead to dramatically different levels of class separation. Representations with higher class separation obtain higher accuracy on the original task, but their features are less useful for downstream tasks. Our results suggest there exists a trade-off between learning invariant features for the original task and features relevant for transfer tasks.
\end{abstract}
\cutsectionup
\section{Introduction}
\cutsectiondown

Features learned by deep neural networks on ImageNet transfer effectively to a wide range of computer vision tasks~\cite{donahue2014decaf,sharif2014cnn,kornblith2019better}.
These networks are often pretrained using vanilla softmax cross-entropy~\cite{bridle1990probabilistic,bridle1990training}, but %
recent work reports that other loss functions such as label smoothing~\cite{szegedy2016rethinking} and sigmoid cross-entropy~\cite{beyer2020we} outperform softmax cross-entropy on ImageNet.
Although a previous investigation suggested that label smoothing and dropout can hurt transfer learning~\cite{kornblith2019better}, the vast majority of work studying alternatives to softmax cross-entropy has restricted its empirical investigation to the accuracy of the trained networks on the original dataset.
While it is valuable to study the impact of objective functions on classification accuracy, %
 improving transfer accuracy of learned representations can have a more significant practical impact for many applications.

This paper takes a comparative approach to understand the effects of different training objectives for image classification through the lens of neural network representations and their transferability.
We carefully tune hyperparameters of each loss function and confirm that several loss functions outperform softmax cross-entropy by a statistically significant margin on ImageNet.
However, we find that these improvements do not transfer to other tasks (see \tabref{tab:transfer} and \figref{fig:imagenet_vs_transfer}).
We delve deeper and explain these empirical findings in terms of effects of different objectives on neural network representations. %
Our key findings are as follows:%
\begin{itemize}[leftmargin=1em,itemsep=0em,topsep=0em]
    \item We analyze the performance of 9 different objectives on ImageNet and in transfer settings. Although many loss functions and regularizers lead to statistically significant improvements over vanilla softmax cross-entropy on ImageNet, these gains do not transfer. These alternative loss functions produce fixed feature extractors that transfer substantially worse to other tasks, in terms of both linear and k-nearest neighbors classification accuracy, and provide no benefit when representations are fully fine-tuned.
    \item The choice of objective primarily affects representations in network layers close to the output. Centered kernel alignment (CKA) reveals large differences in representations of the last few layers of the network, whereas earlier layers are similar regardless of which training objective is used. 
    This helps explain why the choice of objective has little impact on fine-tuning transfer accuracy.
    \item All objectives that improve accuracy over softmax cross-entropy also lead to greater separation between representations of different classes in the penultimate layer features. These alternative objectives appear to collapse within-class variability in representations, which accounts for both the improvement in accuracy on the original task and the reduction in the quality of the features on downstream tasks. %
\end{itemize}
\cutsectionup
\section{Loss Functions and Output Layer Regularizers}
\cutsectiondown
We investigate 9 loss functions and output layer regularizers. %
Let $\bm{z} \in \mathbb{R}^K$ denote the network's output (``logit'') vector, and
let $\bm{t} \in \{0, 1\}^K$ denote a one-hot vector of targets. Let $\bm{x} \in \mathbb{R}^M$ denote the vector of penultimate layer activations, which gives rise to the output vector as $\bm{z} = \bm{W}\bm{x} + \bm{b}$, where $\bm{W} \in \mathbb{R}^{K \times M}$ is the matrix of final layer weights, and $\bm{b}$ is a vector of biases.

All investigated loss functions include a term that encourages $\bm{z}$ to have a high dot product with $\bm{t}$. To avoid solutions that make this dot product large simply by increasing the scale of $\bm{z}$, these loss functions must also include one or more contractive terms and/or normalize $\bm{z}$. Many ``regularizers'' correspond to additional contractive terms added to the loss, so we do not draw a firm distinction between loss functions and regularizers. We describe each objective in detail below, and provide hyperparameters in Appendix~\ref{app:training_details}.

\textbf{Softmax cross-entropy}~\cite{bridle1990probabilistic,bridle1990training} is the de facto loss function for multi-class classification in deep learning:
\begin{align}
\textstyle
\mathcal{L}_\text{softmax}(\bm{z}, \bm{t}) &= -\sum_{k=1}^K t_k \log\left(\frac{e^{z_k}}{\sum_{j=1}^K e^{z_j}}\right) = -\sum_{k=1}^K t_k z_k + \log \sum_{k=1}^K e^{z_k}.
\end{align}
The loss consists of a term that maximizes the dot product between the logits and targets, as well as a contractive term that minimizes the $\mathrm{LogSumExp}$ of the logits. %

\textbf{Label smoothing}~\cite{szegedy2016rethinking} ``smooths'' the targets for softmax cross-entropy. The new targets are given by mixing the original targets with a uniform distribution over all labels, $t' = t \times (1-\alpha) + \alpha / K$, where $\alpha$ determines the weighting of the original and uniform targets. In order to maintain the same scale for the gradient with respect to the positive logit, in our experiments, we scale the label smoothing loss by $1/(1-\alpha)$. The resulting loss is:
\begin{align}
\mathcal{L}_\text{smooth}(\bm{z}, \bm{t}; \alpha) &= -\frac{1}{1-\alpha}\sum_{k=1}^K \left((1-\alpha) t_k + \frac{\alpha}{K}\right) \log\left(\frac{e^{z_k}}{\sum_{j=1}^K e^{z_j}}\right) %
\end{align}
\textcite{muller2019does} previously showed that label smoothing improves calibration and encourages class centroids to lie at the vertices of a regular simplex.

\textbf{Dropout on penultimate layer}: Dropout~\cite{srivastava2014dropout} is among the most prominent regularizers in the deep learning literature. We consider dropout applied to the penultimate layer of the neural network, i.e., when inputs to the final layer are randomly kept with some probability $\rho$. %
When employing dropout, we replace the penultimate layer activations $\bm{x}$ with $\tilde{\bm{x}} = \bm{x} \odot \bm{\xi}/\rho$ where $\xi_i \sim \mathrm{Bernoulli}(\rho)$.
Writing the dropped out logits as $\tilde{\bm{z}} = \bm{W}\tilde{\bm{x}} + \bm{b}$, the dropout loss is:
\begin{align}
    \mathcal{L}_\text{dropout}(\bm{W}, \bm{b}, \bm{x}, \bm{t}; p) = \mathbb{E}_{\bm{\xi}}  \left[\mathcal{L}_\text{softmax}(\tilde{\bm{z}}, \bm{t})\right]
    & = -\sum_{k=1}^K t_k z_k + \mathbb{E}_{\bm{\xi}}\left[\log \sum_{k=1}^K e^{\tilde z_k} \right].
\end{align}
Dropout produces both implicit regularization, by introducing noise into the optimization process~\cite{wei2020implicit}, and explicit regularization, by changing the parameters that minimize the loss~\cite{wager2013dropout}.

\textbf{Extra final layer $\bm{L^2}$ regularization}:
It is common to place the same $L^2$ regularization on the final layer as elsewhere in the network. However, we find that applying greater $L^2$ regularization to the final layer can improve performance. The corresponding loss is:
\begin{align}
\mathcal{L}_\text{extra\_l2}(\bm{W}, \bm{z}, \bm{t}; \lambda_\text{final}) &= \mathcal{L}_\text{softmax}(\bm{z}, \bm{t}) + \lambda_\text{final} \|\bm{W}\|_\mathsf{F}^2.
\end{align}
In architectures with batch normalization, adding additional $L^2$ regularization has no explicit regularizing effect if the learnable scale ($\gamma$) parameters are unregularized, but it still exerts an implicit regularizing effect by altering optimization.

\textbf{Logit penalty}: Whereas label smoothing encourages logits not to be too negative, and dropout imposes a penalty on the logits that depends on the covariance of the weights, an alternative possibility is simply to explicitly constrain logits to be small in $L^2$ norm:
\begin{align}
\mathcal{L}_\text{logit\_penalty}(\bm{z}, \bm{t}; \beta) &= \mathcal{L}_\text{softmax}(\bm{z}, \bm{t}) + \beta \|\bm{z}\|^2.
\end{align}
\citet{dauphin2021deconstructing} showed that this regularizer yields accuracy improvements comparable to dropout.

\textbf{Logit normalization}: We consider the use of $L^2$ \textit{normalization}, rather than regularization, of the logits. Because the entropy of the output of the softmax function depends on the scale of the logits, which is lost after normalization, we introduce an additional temperature parameter $\tau$ that controls the magnitude of the logit vector, and thus, indirectly, the minimum entropy of the output distribution:
\begin{align}
\mathcal{L}_\text{logit\_norm}(\bm{z}, \bm{t}; \tau) = \mathcal{L}_\text{softmax}(\bm{z}/(\tau \|\bm{z}\|), \bm{t}) &= -\frac{1}{\tau\|\bm{z}\|}\sum_{k=1}^K t_k z_k + \log \sum_{k=1}^K e^{z_k/(\tau\|\bm{z}\|)}.
\end{align}
\textbf{Cosine softmax}: We additionally consider $L^2$ normalization of both the penultimate layer features and the final layer weights corresponding to each class. This loss is equivalent to softmax cross-entropy loss if the logits are given by cosine similarity $\mathrm{sim}(\bm{x}, \bm{y}) = \bm{x}^\mathsf{T}\bm{y}/(\|\bm{x}\|\|\bm{y}\|)$ between the weight vector and the penultimate layer plus a per-class bias:
\begin{align}
\mathcal{L}_\text{cos}(\bm{W}, \bm{b}, \bm{x}, \bm{t}; \tau)
=& -\sum_{k=1}^K t_k \left(\mathrm{sim}(\bm{W}_{k,:}, \bm{x})/\tau+ b_k\right) + \log \sum_{k=1}^K e^{\mathrm{sim}(\bm{W}_{k,:}, \bm{x})/\tau + b_k}, \label{eq:cosine_loss}
\end{align}
where $\tau$ is a temperature parameter as above.
Similar losses have appeared in previous literature~\cite{ranjan2017l2,wang2017normface,wojke2018deep,wang2018additive,wang2018cosface,deng2019arcface,liu2017sphereface,chen2018a}, and variants have introduced explicit additive or multiplicative margins to this loss that we do not consider here~\cite{liu2017sphereface,wang2018additive,wang2018cosface,deng2019arcface}. We observe that, even without an explicit margin, manipulating the temperature has a large impact on observed class separation. %

\textbf{Sigmoid cross-entropy}, also known as binary cross-entropy, is the natural analog to softmax cross-entropy for multi-label classification problems. Although we investigate only single-label multi-class classification tasks, we train networks with sigmoid cross-entropy and evaluate accuracy by ranking the logits of the sigmoids. This approach is related to the one-versus-rest strategy for converting binary classifiers to multi-class classifiers. The sigmoid cross-entropy loss is:
\begin{talign}
\mathcal{L}_\text{sce}(\bm{z},\bm{t}) &= -\sum_{k=1}^K \Big(t_k \log\left(\frac{e^{z_k}}{e^{z_k} + 1}\right) + (1 - t_k)\log\left( 1 - \frac{e^{z_k}}{e^{z_k} + 1}\right)\Big)\nonumber\\
&= -\sum_{k=1}^K t_k z_k + \sum_{k=1}^K \log (e^{z_k} + 1).
\end{talign}
The $\mathrm{LogSumExp}$ term of softmax loss is replaced with the sum of the $\mathrm{softplus}$-transformed logits. We initialize the biases of the logits $\bm{b}$ to $-\log(K)$ so that the initial output probabilities are approximately $1/K$. \citet{beyer2020we} have previously shown that sigmoid cross-entropy loss leads to improved accuracy on ImageNet relative to softmax cross-entropy.

\textbf{Squared error}: Finally, we investigate squared error loss, as formulated by \textcite{hui2020evaluation}:
\begin{align}
\mathcal{L}_\text{mse}(\bm{z}, \bm{t}; \kappa, M) &= \frac{1}{K}\sum_{k=1}^K  \left(\kappa t_k (z_k - M)^2 + (1-t_k) z_k^2\right),\
\end{align}
where $\kappa$ and $M$ are hyperparameters. $\kappa$ sets the strength of the loss for the correct class relative to incorrect classes, whereas $M$ controls the magnitude of the correct class target. When $\kappa = M = 1$, the loss is simply the mean squared error between $\bm{z}$ and $\bm{t}$. Like \textcite{hui2020evaluation}, we find that placing greater weight on the correct class slightly improves ImageNet accuracy. %

\cutsectionup
\section{Results}
\cutsectiondown
\vspace{-0.5em}

For each loss, we trained 8 ResNet-50~\cite{he2016deep,resnetv15torchblogpost} models on the ImageNet ILSVRC 2012 dataset~\cite{deng2009imagenet,russakovsky2015imagenet}. To tune loss hyperparameters and the epoch for early stopping, we performed 3 training runs per hyperparameter configuration where we held out a validation set of 50,046 ImageNet training examples. %
We provide further details regarding the experimental setup in Appendix~\ref{app:all_training_details}. We also confirm that our main findings hold for Inception v3~\cite{szegedy2016rethinking} models in Appendix~\ref{app:inception}.

\subsection{Better objectives improve accuracy, but do not transfer better}
\vspace{-0.5em}
\begin{table*}[h]
  \caption{\textbf{Objectives that produce higher ImageNet accuracy lead to less transferable fixed features.}  ``ImageNet'' columns reflect accuracy of ResNet-50 models on the ImageNet validation set. ``Transfer'' columns reflect accuracy of $L^2$-regularized multinomial logistic regression or $k$-nearest neighbors classifiers trained to classify different transfer datasets using the fixed penultimate layer features of the ImageNet-trained networks. Numbers are averaged over 8 different pretraining runs; values not significantly different than the best are bold-faced ($p < 0.05$, $t$-test). The strength of $L^2$ regularization is selected on a holdout set, and $k$ is selected using leave-one-out cross-validation on the training set. %
  See Appendix Table~\ref{tab:transfer_inception} for a similar table for Inception v3 models, Appendix~\ref{app:training_dynamics_transfer} for linear transfer accuracy evaluated at different epochs of ImageNet pretraining, and Appendix~\ref{app:chexpert} for similar findings on the Chexpert chest X-ray dataset~\cite{irvin2019chexpert}.}
  \aftertablecaption
\vspace{-0.4em}
  \centering
  \fontsize{8}{10}\selectfont
  \setlength{\tabcolsep}{0.5em}
\begin{tabular}{lrrrrrrrrrr}
\toprule
\multicolumn{1}{c}{} & \multicolumn{2}{c}{ImageNet}  & \multicolumn{8}{c}{Transfer}
\\
\cmidrule(lr){2-3}
\cmidrule(lr){4-11}
\multicolumn{1}{c}{Pretraining loss} & \multicolumn{1}{c}{Top-1} & \multicolumn{1}{c}{Top-5} &         Food &         CIFAR10 &        CIFAR100 &        Birdsnap &          SUN397 &   Cars &      Pets & \multicolumn{1}{c}{Flowers}\\
\midrule
\multicolumn{10}{l}{\textit{Linear transfer:}}\\
Softmax             & 77.0 & 93.40 &  \textbf{74.6} &  \textbf{92.4} &  \textbf{76.9} &  \textbf{55.4} &  \textbf{62.0} &  \textbf{60.3} &           92.0 &  \textbf{94.0}            \\
Label smoothing     & 77.6 & 93.78 &           72.7 &           91.6 &           75.2 &           53.6 &           61.6 &           54.8 &  \textbf{92.9} &           91.9            \\
Sigmoid             & \textbf{77.9} & 93.50 &           73.4 &           91.7 &           75.7 &           52.3 &  \textbf{62.0} &           56.1 &           92.5 &           92.9            \\
More final layer $L^2$         & 77.7 & 93.79 &           70.6 &           91.0 &           73.7 &           51.5 &           60.1 &           50.3 &           92.4 &           89.8           \\
Dropout             & 77.5 & 93.62             &           72.6 &           91.4 &           75.0 &           53.6 &           61.2 &           54.7 &  \textbf{92.6} &           92.1            \\
Logit penalty       & 77.7 & \textbf{93.83} &           68.1 &           90.2 &           72.3 &           48.1 &           59.0 &           48.3 &           92.3 &           86.6            \\
Logit normalization          & \textbf{77.8} & 93.71 &   66.3 &           90.5 &           72.9 &           50.7 &           58.1 &           45.4 &           92.0 &           82.9           \\
Cosine softmax      & \textbf{77.9} & \textbf{93.86} &           62.0 &           89.9 &           71.3 &           45.4 &           55.0 &           36.7 &           91.1 &           75.3\\ %
Squared error       & 77.2 & 92.79 &           39.8 &           82.2 &           56.3 &           21.8 &           39.9 &           15.3 &           84.7 &           46.7\\ %
\midrule
\multicolumn{10}{l}{\textit{K-nearest neighbors:}}\\
Softmax             & 77.0 & 93.40 &  \textbf{60.9} &  \textbf{88.8} &  \textbf{67.4} &  \textbf{38.4} &           53.0 &  \textbf{28.9} &           88.8 &  \textbf{83.6} \\
Label smoothing     & 77.6 & 93.78 &           59.2 &           88.3 &           66.3 &  \textbf{39.2} &  \textbf{53.5} &           27.3 &  \textbf{91.5} &           80.3 \\
Sigmoid             & \textbf{77.9} & 93.50  &           58.5 &           88.2 &           66.6 &           34.4 &           52.8 &           26.7 &           90.8 &           81.5 \\
More final layer $L^2$         & 77.7 & 93.79 &           55.0 &           87.7 &           65.0 &           36.2 &           51.1 &           24.4 &           90.9 &           75.6 \\
Dropout             & 77.5 & 93.62         &           58.4 &           88.2 &           66.4 &           38.1 &           52.4 &           27.8 &           91.1 &           80.1 \\
Logit penalty       & 77.7 & \textbf{93.83}   &           52.5 &           86.5 &           62.9 &           31.0 &           49.3 &           22.4 &           90.8 &           68.1 \\
Logit normalization          & \textbf{77.8} & 93.71         &           50.9 &           86.6 &           63.1 &           35.1 &           45.8 &           24.1 &           88.2 &           63.1 \\
Cosine softmax      & \textbf{77.9} & \textbf{93.86} &           45.5 &           86.0 &           61.5 &           28.8 &           41.8 &           19.0 &           87.0 &           52.7 \\
Squared error       & 77.2 & 92.79   &           27.7 &           74.5 &           44.3 &           13.8 &           28.6 &            9.1 &           82.6 &           28.2 \\
\bottomrule
\end{tabular}
  \label{tab:transfer}
\vspace{-0.3em}
\end{table*}

\begin{wrapfigure}{r}{0.33\textwidth}
  \begin{center}
     \vskip -1.5em
    \includegraphics[width=0.3125\textwidth]{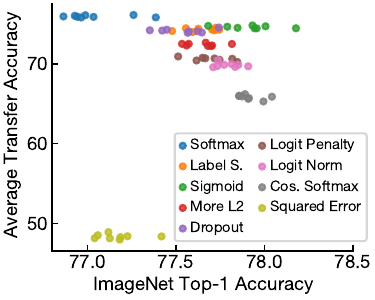}
  \end{center}
  \vskip -0.5em
  \caption{\textbf{Higher ImageNet accuracy is not associated with higher linear transfer accuracy.} Points represent individual training runs. See Appendix Figure~\ref{fig:imagenet_vs_transfer_datasets} for similar plots for individual datasets.}
  \label{fig:imagenet_vs_transfer}
  \vskip -1.5em
\end{wrapfigure}
We found that, when properly tuned, all investigated objectives except squared error provide a statistically significant improvement over softmax cross-entropy, as shown in the left two columns of Table~\ref{tab:transfer}. The gains are small but meaningful, with sigmoid cross-entropy and cosine softmax both leading to an improvement of 0.9\% in top-1 accuracy over the baseline. For further discussion of differences in the training curves, robustness, calibration, and predictions of these models, see Appendix~\ref{app:other_models}.

Although networks trained with softmax cross-entropy attain lower ImageNet top-1 accuracy than any other loss function, they nonetheless provide the most transferable features. We evaluated the transferability of the fixed features of our ImageNet-pretrained models by training linear or k-nearest neighbors (kNN) classifiers to classify 8 different natural image datasets: Food-101~\citep{bossard2014food}, CIFAR-10 and CIFAR-100~\citep{krizhevsky2009learning}, Birdsnap~\citep{berg2014birdsnap}, SUN397~\citep{xiao2010sun}, Stanford Cars~\citep{krause2013collecting}, Oxford-IIIT Pets~\citep{parkhi2012cats}, and Oxford Flowers~\citep{nilsback2008automated}. The results of these experiments are shown in Table~\ref{tab:transfer} and  Figure~\ref{fig:imagenet_vs_transfer}. In both linear and kNN settings, representations learned with vanilla softmax cross-entropy perform best for most tasks.

As shown in Table~\ref{tab:finetune}, when networks are fully fine-tuned on downstream tasks, the pretraining objective has little effect on the resulting accuracy. When averaging across all tasks, the best loss provides only a 0.2\% improvement over softmax cross-entropy, which does not reach statistical significance and is much smaller than the 0.9\% difference in ImageNet top-1 accuracy. Thus, using a different loss function for pretraining can improve accuracy on the pretraining task, but this improvement does not appear to transfer to downstream tasks.

\begin{table*}
  \caption{\textbf{The training objective has little impact on the performance of fine-tuned networks.} Accuracy of fine-tuning pretrained networks on transfer datasets, averaged over 3 different pretraining initializations. We tune hyperparameters separately for each objective and dataset. Numbers not significantly different than the best are bold-faced ($p < 0.05$, $t$-test for individual datasets, two-way ANOVA for average). See Appendix~\ref{app:fine_tuning} for training details.}
  \aftertablecaption
  \centering
  \footnotesize
  \setlength{\tabcolsep}{0.45em}
\begin{tabular}{lrrrrrrrrr}
\toprule
Pretraining loss &    Food & CIFAR10 & CIFAR100 & Birdsnap &  SUN397 &    Cars &    Pets & Flowers & Avg. \\
\midrule
Softmax             &  \textbf{88.2} &  \textbf{96.9} &  \textbf{84.1} &  \textbf{76.2} &           63.4 &           91.3 &           93.1 &  \textbf{96.7} & \textbf{86.2} \\
Label smoothing     &  \textbf{88.3} &  \textbf{96.7} &           84.0 &  \textbf{76.3} &           63.6 &           91.2 &  \textbf{93.7} &  \textbf{96.3} & \textbf{86.3} \\
Sigmoid             &  \textbf{88.3} &  \textbf{96.9} &           83.6 &  \textbf{76.2} &           63.5 &  \textbf{91.8} &  \textbf{93.7} &  \textbf{96.3} & \textbf{86.3} \\
More final layer $L^2$ &           88.1 &  \textbf{96.9} &  \textbf{84.4} &           75.9 &  \textbf{64.5} &  \textbf{91.5} &  \textbf{93.8} &  \textbf{96.2} & \textbf{86.4} \\
Dropout             &  \textbf{88.3} &  \textbf{96.7} &  \textbf{84.2} &  \textbf{76.5} &           63.9 &           91.2 &  \textbf{93.9} &  \textbf{96.3} & \textbf{86.4} \\
Logit penalty       &  \textbf{88.4} &  \textbf{96.9} &           83.9 &  \textbf{76.4} &           63.3 &           91.2 &  \textbf{93.4} &  \textbf{96.0} & 86.2 \\
Logit normalization &           87.9 &  \textbf{96.9} &           82.9 &  \textbf{76.0} &           58.3 &           91.2 &           92.7 &           95.9 & 85.2 \\
Cosine softmax      &  \textbf{88.3} &  \textbf{96.9} &           83.2 &  \textbf{75.6} &           56.9 &           91.3 &           92.5 &           95.9 & 85.1 \\
Squared error       &           87.8 &  \textbf{96.9} &  \textbf{84.0} &           75.7 &           61.0 &           91.4 &           93.0 &           95.2 & 85.6\\
\bottomrule
\end{tabular}
  \label{tab:finetune}
  \vskip -1em
\end{table*}

\cutsubsectionup
\subsection{The choice of objective primarily affects hidden representations close to the output}
\cutsubsectiondown
\label{sec:similarity}

\begin{figure*}[t!]
\begin{center}
      \includegraphics[width=\textwidth]{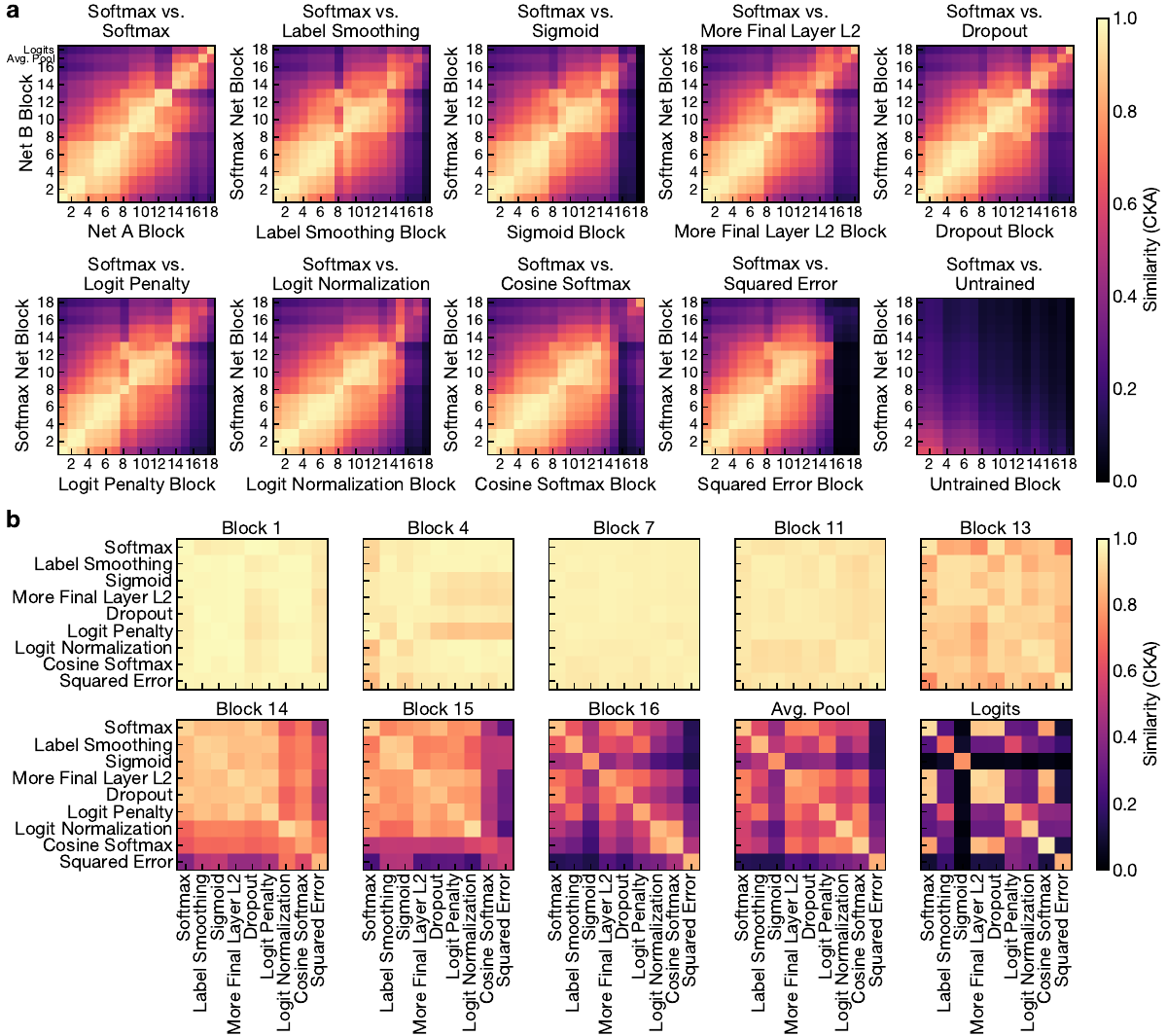}
\end{center}
\vskip -0.6em
\caption{\textbf{The choice of loss function affects representations only in later network layers.} All plots show linear centered kernel alignment (CKA) between representations computed on the ImageNet validation set. \textbf{a}: CKA between network layers, for pairs of ResNet-50 models trained from different initializations with different losses. As controls, the top-right plot shows CKA between two networks trained with softmax, and the bottom-right plot shows CKA between a model trained with softmax loss and a model at initialization where batch norm moments have been computed on the training set. \textbf{b}: CKA between representations extracted from corresponding layers of networks trained with different loss functions. Diagonal reflects similarity of networks with the same loss function trained from different initalizations. See Appendix Figure~\ref{fig:cka_inception} for a similar figure for Inception v3 models.}
\label{fig:cka}
\vskip -0.5em
\end{figure*}

\begin{figure*}[t!]
\begin{center}
      \includegraphics[width=\textwidth]{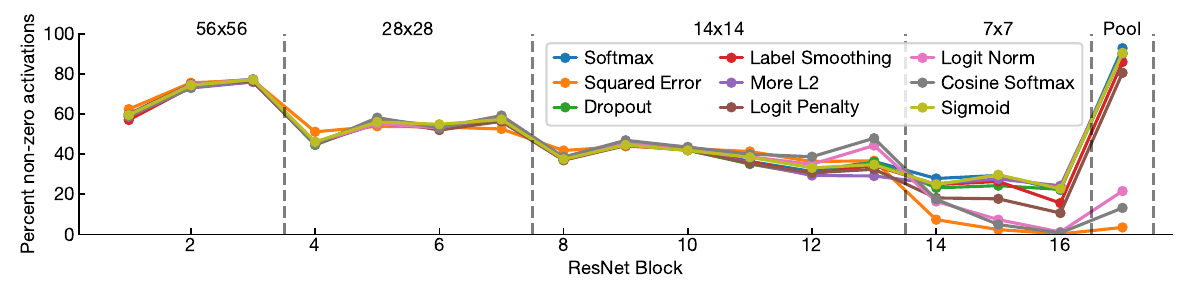}
\end{center}
\vskip -1.5em
\caption{\textbf{Loss functions affect sparsity of later layer representations.} Plot shows the average \% non-zero activations for each ResNet-50 block, after the residual connection and subsequent nonlinearity, on the ImageNet validation set. Dashed lines indicate boundaries between stages.}
\label{fig:sparsity}
\vspace*{-1em}
\end{figure*}

Our observation that ``improved'' objectives improve only on ImageNet and not on transfer tasks raises questions about which representations, exactly, they affect.
We use two tools to investigate differences in the hidden representations of networks trained with different loss functions. First, we use centered kernel alignment~\cite{kornblith2019similarity,cortes2012algorithms,cristianini2002kernel} to directly measure the similarity of hidden representations across networks trained with different loss functions. Second, we measure the sparsity of the ReLU activations in each layer. Both analyses suggest that all loss functions learn similar representations throughout the majority of the network, and differences are present only in the last few ResNet blocks.

Linear centered kernel alignment (CKA) provides a way to measure similarity of neural network representations that is invariant to rotation and isotropic scaling in representation space~\cite{kornblith2019similarity,cortes2012algorithms,cristianini2002kernel}. Unlike other ways of measuring representational similarity between neural networks, linear CKA can identify architectural correspondences between layers of networks trained from different initializations~\cite{kornblith2019similarity}, a prerequisite for comparing networks trained with different objectives. Given two matrices $\bm{X} \in \mathbb{R}^{n \times p_1}$ and $\bm{Y} \in \mathbb{R}^{n \times p_2}$ containing activations to the same $n$ examples, linear CKA computes the cosine similarity between the reshaped $n\times n$ covariance matrices between examples:%
\begin{align}
    \mathrm{CKA}_\text{linear}(\bm{X}, \bm{Y}) &= \frac{\mathrm{vec}(\mathrm{cov}(\bm{X})) \cdot \mathrm{vec}(\mathrm{cov}(\bm{Y}))}{\|\mathrm{cov}(\bm{X})\|_\text{F}\|\mathrm{cov}(\bm{Y})\|_\text{F}}.
\end{align}
We measured CKA between all possible pairings of ResNet blocks from 18 networks (2 different initializations for each objective). To reduce memory requirements, we used minibatch CKA~\cite{nguyen2021do} with minibatches of size 1500 and processed the ImageNet validation set for 10 epochs.

As shown in Figure~\ref{fig:cka}, representations of the majority of network layers are highly similar regardless of loss function, but late layers differ substantially. Figure~\ref{fig:cka}a shows similarity between all pairs of blocks from pairs of networks, where one network is trained with vanilla softmax cross-entropy and the other is trained with either softmax or a different loss function. The diagonals of these plots indicate the similarity between architecturally corresponding layers. For all network pairs, the diagonals are stronger than the off-diagonals, indicating that architecturally corresponding layers are more similar than non-corresponding layers. However, in the last few layers of the network, the diagonals are substantially brighter when both networks are trained with softmax than when the second network is trained with a different loss, indicating representational differences in these layers. Figure~\ref{fig:cka}b shows similarity of representations among all loss functions for a subset of blocks. Consistent differences among loss functions are present only in the last third of the network, starting around block 13. %

The sparsity of activations reveals a similar pattern of layer-wise differences between networks trained with different loss functions.  Figure~\ref{fig:sparsity} shows the proportion of non-zero activations in different layers. In all networks, the percentage of non-zero ReLU activations decreases with depth, attaining its minimum at the last convolutional layer. In the first three ResNet stages, activation sparsity is broadly similar regardless of the loss. However, in the final stage and penultimate average pooling layer, the degree of sparsity depends greatly on the loss.
Penultimate layer representations of vanilla softmax networks are the least sparse, with 92.8\% non-zero activations. Logit normalization, cosine softmax, and squared error all result in much sparser ($<$25\% non-zero) activations.

These results immediately suggest an explanation for the limited effect of the training objective on networks' fine-tuning performance. We observe differences among objectives only in later network layers, but previous work has found that these layers change substantially during fine-tuning~\cite{yosinski2014how,raghu2019transfusion,neyshabur2020what}, an observation we replicate in Appendix~\ref{app:cka_transfer}. Thus, the choice of training objective appears to affect parts of the network that are specific to the pretraining task, and do not transfer when the network is fine-tuned on other tasks.

\cutsubsectionup
\subsection{Regularization and alternative losses increase class separation}
\label{sec:separation}
\cutsubsectiondown

The previous section suggests that different loss functions learn very different penultimate layer representations, even when their overall accuracy is similar. However, as shown in Appendix~\ref{app:combined_losses}, combining different losses and penultimate layer regularizers provides no accuracy improvements over training with only one, suggesting that they share similar mechanisms. In this section, we demonstrate that a simple property of networks' penultimate layer representations can explain their beneficial effect on accuracy relative to vanilla softmax cross-entropy, as well as their harmful effect on fixed features.
Specifically, compared to vanilla softmax cross-entropy, all investigated losses cause the network to reduce the relative within-class variance in the penultimate layer representation space. This reduction in within-class variance corresponds to increased separation between classes, and is harmful to linear transfer.   %

\begin{figure*}
\begin{minipage}[l]{0.48\linewidth}
\captionof{table}{\textbf{Regularization and alternative losses improve class separation in the penultimate layer.} Results are averaged over 8 ResNet-50 models per loss on the ImageNet training set.}
\centering
\label{tab:r2}
  \aftertablecaption
\footnotesize
  \setlength{\tabcolsep}{0.35em}
\vspace{-0.3em}
\begin{tabular}{lrr}
    \toprule
    Loss/regularizer & \multicolumn{1}{c}{\centering ImageNet top-1} & \multicolumn{1}{c}{\centering Class sep. ($R^2$)} \\
    \midrule
    Softmax & 77.0 $\pm$ 0.06 & 0.349 $\pm$ 0.0002    \\
    Label smooth. & 77.6 $\pm$ 0.03 & 0.420 $\pm$ 0.0003\\
    Sigmoid  & 77.9 $\pm$ 0.05 & 0.427  $\pm$ 0.0003\\
    Extra $L^2$  & 77.7 $\pm$ 0.03 & 0.572  $\pm$ 0.0006\\
    Dropout  & 77.5 $\pm$ 0.04 & 0.461 $\pm$ 0.0003 \\
    Logit penalty  & 77.7 $\pm$ 0.04 & 0.601  $\pm$ 0.0004\\
    Logit norm  & 77.8 $\pm$ 0.02 & 0.517  $\pm$ 0.0002\\
    Cosine softmax  & 77.9 $\pm$ 0.02 & 0.641  $\pm$ 0.0003\\
    Squared error  & 77.2 $\pm$ 0.04 & 0.845  $\pm$ 0.0002\\
    \bottomrule
\end{tabular}
\end{minipage}
\hfill
\begin{minipage}[r]{0.5\linewidth}
\begin{center}
      \includegraphics[width=\linewidth]{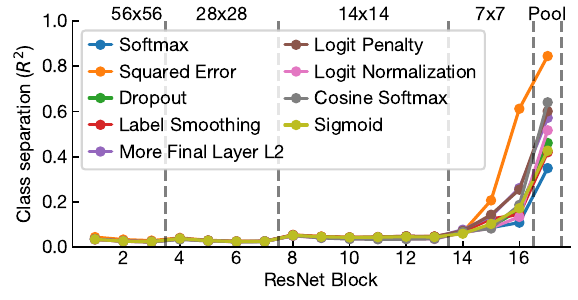}
\end{center}
\vskip -1em
\caption{Class separation in different layers of ResNet-50, averaged over 8 models per loss on the ImageNet training set. For convolutional layers, we compute cosine distances by flattening the representations of examples across spatial dimensions.}
\label{fig:r2}
\end{minipage}\hfill
\vspace{-0.75em}
\end{figure*}

The ratio of the average within-class cosine distance to the overall average cosine distance measures the dispersion of representations of examples belonging to the same class relative to the overall dispersion of embeddings. We take one minus this quantity to get a closed-form index of class separation that is between 0 and 1:
\begin{samepage}
\begin{align}
    R^2 = 1 - \bar{d}_\text{within}/\bar{d}_\text{total}\label{eq:class_separation}
\end{align}
\vskip -2em
\begin{align*}
    \bar{d}_\text{within} = \sum_{k=1}^K \sum_{m=1}^{N_k} \sum_{n=1}^{N_k} \frac{1 - \mathrm{sim}(\bm{x}_{k,m}, \bm{x}_{k,n})}{KN_k^2}, &&
    \bar{d}_\text{total} = \sum_{j=1}^K\sum_{k=1}^K\sum_{m=1}^{N_j} \sum_{n=1}^{N_k} \frac{1 - \mathrm{sim}(\bm{x}_{j,m}, \bm{x}_{k,n})}{K^2 N_j N_k}
\end{align*}
\end{samepage}
where $\bm{x}_{k, m}$ is the embedding of example $m$ in class $k \in \{1, \ldots, K\}$, $N_k$ is the number of examples in class $k$, and $\mathrm{sim}(\bm{x}, \bm{y}) = \bm{x}^\mathsf{T}\bm{y}/(\|\bm{x}\|\|\bm{y}\|)$ is cosine similarity between vectors. As we show in Appendix~\ref{app:relationships}, if the embeddings are first $L^2$ normalized, then $1-R^2$ is the ratio of the average within-class variance to the weighted total variance, where the weights are inversely proportional to the number of examples in each class. For a balanced dataset, $R^2$ is equivalent to centered kernel alignment~\cite{cortes2012algorithms,cristianini2002kernel} between the embeddings and the one-hot label matrix, with a cosine kernel.  See Appendix~\ref{app:other_ratios} for results with other distance metrics.

As shown in Table~\ref{tab:r2} and Figure~\ref{fig:r2}, all regularizers and alternative loss functions we investigate produce greater class separation in penultimate layer representations as compared to vanilla softmax loss. Importantly, this increase in class separation is specific to forms of regularization that affect networks' final layers. In Appendix~\ref{app:autoaugment_class_separation}, we show that adding regularization through data augmentation improves accuracy without a substantial change in class separation. %
We further investigate the training dynamics of class separation in Appendix~\ref{app:training_dynamics}, finding that, for softmax cross-entropy, class separation peaks early in training and then falls, whereas for other objectives, class separation either saturates or continues to rise as training progresses. %

\subsection{Greater class separation is associated with less transferable features}

\begin{figure*}[t]
\begin{minipage}[l]{0.28\linewidth}
\caption{\textbf{Class separation negatively correlates with linear transfer accuracy.} \textbf{a}: Mean linear transfer accuracy across all tasks vs. class separation on ImageNet, for different loss functions. \textbf{b}: For cosine softmax loss, training at higher temperature produces greater class separation but worse linear transfer accuracy.}
\label{fig:scatter}
\end{minipage}
\hfill
\begin{minipage}[l]{0.6875\linewidth}
\begin{center}
      \includegraphics[width=\textwidth]{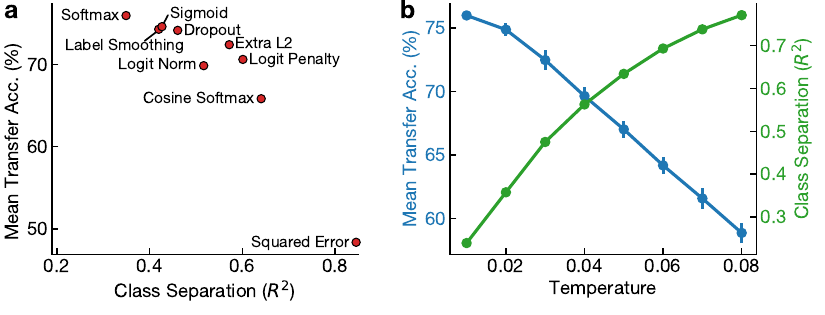}
\end{center}
\end{minipage}
\vskip -1em
\end{figure*}

Although losses that improve class separation lead to higher accuracy on the ImageNet validation set, the feature extractors they learn transfer worse to other tasks. Figure~\ref{fig:scatter}a plots mean linear transfer accuracy versus class separation for each of the losses we investigate. We observe a significant negative correlation (Spearman's $\rho = -0.93$, $p = 0.002$). Notably, vanilla softmax cross-entropy produces the least class separation and the most transferable features, whereas squared error produces much greater class separation than other losses and leads to much lower transfer performance.

To confirm this relationship between class separation, ImageNet accuracy, and transfer, we trained models with cosine softmax with varying values of the temperature parameter $\tau$.\footnote{Training at $\tau < 0.05$ was unstable; we scale the loss by the $\tau$ to reduce the effect of $\tau$ on the size of the gradient WRT the correct class logit. Relationships for $\tau \geq$ 0.05 remain consistent without loss scaling.} As shown in Table~\ref{tab:temperature}, lower temperatures yield lower top-1 accuracies and worse class separation. However, even though the lowest temperature of $\tau = 0.01$ achieved 2.7\% lower accuracy on ImageNet than the best temperature, this lowest temperature gave the best features for nearly all transfer datasets. Thus, $\tau$ controls a tradeoff between the generalizability of penultimate-layer features and the accuracy on the target dataset. For $\tau \geq 0.04$, ImageNet accuracy saturates, but, as shown in Figure~\ref{fig:scatter}b, class separation continues to increase and transfer accuracy continues to decrease.

\begin{table*}[t]
  \caption{\textbf{Temperature of cosine softmax loss controls ImageNet accuracy, class separation ($R^2$), and linear transfer accuracy.}}
  \aftertablecaption
  \centering
  \footnotesize
  \setlength{\tabcolsep}{0.55em}
\begin{tabular}{lrrrrrrrrrr}
\toprule
\multicolumn{1}{c}{} & \multicolumn{2}{c}{ImageNet} & \multicolumn{8}{c}{Transfer}
\\
\cmidrule(lr){2-3}
\cmidrule(lr){4-11}
\multicolumn{1}{c}{Temp.} & \multicolumn{1}{c}{Top-1} & \multicolumn{1}{c}{$R^2$} &         Food &         CIFAR10 &        CIFAR100 &        Birdsnap &          SUN397 &   Cars &      Pets & \multicolumn{1}{c}{Flowers} \\%

\midrule
0.01 & \textcolor{red}{74.9} & \textcolor{red}{0.236} &    \textbf{73.4} &    \textbf{91.9} &     \textbf{76.5} &     \textbf{57.2} &   \textbf{60.5} &          \textbf{62.9} &        91.7 &              \textbf{93.6}  \\
0.02 & 77.0 & 0.358 &   72.1 &    91.8 &     76.2 &     56.5 &   60.4 &          58.5 &        92.2 &              91.2 \\ %
0.03 & 77.5 & 0.475 &   69.1 &    91.5 &     74.9 &     53.7 &   59.1 &          51.8 &        \textbf{92.3} &              87.4 \\ %
0.04 & \textbf{77.6} & 0.562 &   66.0 &    90.7 &     73.8 &     50.3 &   57.4 &          45.1 &        91.7 &              82.2 \\ %
0.05 & \textbf{77.6} & 0.634 &   62.8 &    90.4 &     72.2 &     47.6 &   55.4 &          38.6 &        91.0 &              78.3 \\ %
0.06 & 77.5 & 0.693 &   60.3 &    89.3 &     69.8 &     43.3 &   53.8 &          33.3 &        91.0 &              72.7 \\
0.07 & 77.5 & 0.738 &   57.1 &    88.7 &     68.6 &     39.6 &   51.4 &          29.1 &        90.2 &              67.9 \\
0.08 & \textbf{77.6} & \textbf{0.770} &   53.7 &    87.7 &     66.5 &     35.5 &   49.4 &          25.7 &        89.3 &              63.2 \\
\bottomrule
\end{tabular}
  \label{tab:temperature}
\end{table*}

\begin{figure*}[t]
\begin{minipage}[l]{0.6\linewidth}

\caption{\textbf{Class separation positively correlates with accuracy when relearning how to classify ImageNet classes from limited data.} All accuracy numbers are computed on a 10,000 example subset of the ImageNet validation set (10 examples per class). Blue dots indicate accuracy of the weights of the original ImageNet-trained model. Orange dots indicate accuracy of a linear classifier trained on the other 40,000 examples from the ImageNet validation set. Lines connect blue and orange dots corresponding to the same objective. The gap between the accuracies of the original and relearned classifiers narrows as class separation increases. Class separation is measured on the ImageNet training set. All numbers are averaged over 8 models.}
\label{fig:class_sep_imagenet}
\end{minipage}
\hfill
\begin{minipage}[l]{0.375\linewidth}
\begin{center}
      \includegraphics[width=\textwidth]{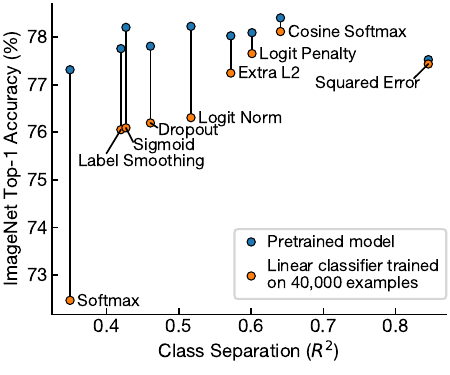}
\end{center}
\end{minipage}
\vskip -1.3em
\end{figure*}

Is there any situation where features with greater class separation could be beneficial for a downstream task? In Figure~\ref{fig:class_sep_imagenet}, we use $L^2$-regularized logistic regression to relearn the original 1000-way ImageNet classification head from penultimate layer representations of 40,000 examples from the ImageNet validation set.\footnote{We use this experimental setup to avoid training linear classifiers on examples that were also seen during pretraining. However, results are qualitatively similar if we train the linear classifier on a 50,046 example subset of the training set and test on the full validation set.} We find that features from networks trained with vanilla softmax loss perform \textit{worst}, whereas features from networks with greater class separation perform substantially better.
Thus, it seems that representations with greater class separation are ``overfit,'' not to the pretraining datapoints, but to the pretraining \textit{classes}---they perform better for classifying these classes, but worse when the downstream task requires classifying different classes.

Our results above provide some further evidence that greater class separation can be beneficial in real-world scenarios where downstream datasets share classes with the pretraining dataset. In Tables~\ref{tab:transfer} and \ref{tab:temperature}, representations with the lowest class separation perform best on all datasets except for Oxford-IIIT Pets, where representations with slightly greater class separation consistently perform slightly better. We thus explore class overlap between transfer datasets and ImageNet in Appendix~\ref{app:class_level_overlap}, and find that, of the 37 cat and dog breeds in Oxford-IIIT Pets, 25 correspond directly to ImageNet classes. Furthermore, it is possible to achieve 71.2\% accuracy on Oxford-IIIT Pets simply by taking the top-1 predictions of an ImageNet classifier and and mapping them directly to its classes, but this strategy achieves much lower accuracy on other datasets. %

\cutsectionup
\vskip -0.5em
\section{Related work}
\cutsectiondown

\textbf{Understanding training objectives.}
There is a great deal of previous work that investigates why some objectives perform better than others, using both theoretical and empirical approaches. On linearly separable data, theoretical analysis shows that gradient descent on both unregularized logistic or multinomial logistic regression objectives (i.e., linear models with sigmoid or softmax cross-entropy loss) eventually converges to the minimum norm solution~\cite{soudry2018implicit}. These results can be extended to neural networks in certain restricted settings~\cite{gunasekar2018implicit,wei2019regularization}. However, the convergence to this solution is very slow. Theoretical analysis of dropout has bounded the excess risk of single-layer~\cite{wager2014altitude} and two-layer~\cite{mianjy2020convergence} networks. Empirical studies have attempted to understand regularization in more realistic settings. Label smoothing has been explained in terms of mitigation of label noise~\cite{lukasik2020does,chen2020investigation}, entropy regularization~\cite{pereyra2017regularizing,meister2020generalized}, and accelerated convergence~\cite{xu2020towards}. Other work has shown that dropout on intermediate layers has both implicit and explicit effects, and both are important to accuracy~\cite{wei2019regularization}.
\citet{janocha2016loss} previously studied both empirical and mathematical similarities and differences among loss functions for supervised classification in the context of deep learning.

\textbf{Training objectives for transfer.}
Our study of the transferability of networks trained with different objectives extends the previous investigation of~\citet{kornblith2019better}, which showed that two of the regularizers considered here (label smoothing and dropout) lead to features that transfer worse. Similar results have been reported for self-supervised learning, where the loss parameters that maximize accuracy on the contrastive task do not provide the best features for linear classification~\cite{chen2020simple}.
Our work demonstrates that this phenomenon is pervasive, and connects it to properties of hidden representations. By contrast, work that explores out-of-distribution (OOD) generalization, where the training and evaluation datasets consist of the same classes but with some degree of distribution shift, finds that ImageNet accuracy is highly predictive of OOD accuracy across many pretrained models~\cite{recht2019imagenet,taori2020measuring}. In Appendix Table~\ref{tab:ood}, we show that, across loss functions, ImageNet validation set accuracy is not entirely correlated with OOD accuracy, but different losses produce much smaller differences in OOD accuracy than linear transfer accuracy.

Other related work has attempted to devise loss functions to learn more transferable embeddings for few-shot learning~\cite{snell2017prototypical,sung2018learning,bertinetto2018metalearning,chen2018a,oreshkin2018tadam}, but embeddings of models trained with softmax cross-entropy often perform on par with these more sophisticated techniques~\cite{tian2020rethinking}. \citet{doersch2020crosstransformers} motivate their few-shot learning method as a way to mitigate ``supervision collapse,'' which they do not directly quantify, but is closely related to our notion of class separation. While our paper was under review, two studies reported advantages for increased final layer $L^2$ regularization when performing few-shot linear transfer from very large pretraining datasets to some, but not all, downstream tasks~\cite{zhai2021scaling,abnar2021exploring}. Our findings suggest that these advantages may arise because the pretraining and downstream tasks share classes, but further investigation is needed to confirm this hypothesis.

\textbf{Class separation.} 
Prior work has investigated class separation in neural networks in a variety of different ways.  Theoretical work connects various concepts of the normalized margin between classes to the generalization properties of neural networks~\cite{bartlett1997valid,neyshabur2015norm,bartlett2017spectrally,neyshabur2018pac}. Empirically, \citet{chen2019angular} measure class separation via ``angular visual hardness,'' the arccosine-transformed cosine similarity between the weight vectors and examples, and suggested an association between this metric and accuracy. However, as shown in Appendix Figure~\ref{fig:avh}, this metric is unable to differentiate between networks trained with softmax and sigmoid cross-entropy. \citet{papyan2020prevalence} describe a phenomenon they call ``neural collapse,'' which occurs after training error vanishes and is associated with the collapse of the class means, classifier, and activations to the vertices of an equiangular tight frame, implying maximal class separation. We relate our observations to theirs in Appendix Table~\ref{tab:simplex_etf}.  More generally, studies have investigated how class information evolves through the hidden layers of neural networks using linear classifiers~\cite{alain2016understanding}, binning estimators of mutual information~\cite{shwartz2017opening,saxe2019information,goldfeld2018estimating}, Euclidean distances~\cite{schilling2018deep}, the soft nearest neighbor loss~\cite{frosst2019analyzing}, and manifold geometry~\cite{ansuini2019intrinsic,cohen2020separability}. %
In the context of a study of representational consistency across networks trained from different initializations, \citet{mehrer2020individual} previously reported increased class separation in CIFAR-10 networks trained with dropout.

Other work has explored relationships between measures of class separation in embedding spaces and accuracy in few-shot and deep metric learning settings. In deep metric learning, \citet{roth2020revisiting} show that objectives that lead to greater class separation generally produce higher recall, and find a similar relationship for a measure of the uniformity of the singular value spectrum of the representation space. In few-shot learning, different work has reached different conclusions: \citet{goldblum2020unraveling} find that regularization that \textit{increases} class separation improves few-shot classification performance on mini-ImageNet and CIFAR-FS, whereas \citet{liu2020negative} find that introducing a negative margin into a cosine softmax loss yields both lower class separation and higher accuracy.

\vspace{-0.35em}
\section{Limitations}
\label{sec:limitations}
\vspace{-0.35em}
Although we investigate many losses and multiple architectures, our experiments are limited to moderately sized datasets with moderately sized models, and our conclusions are limited to supervised classification settings. ResNet-50 on trained on ImageNet is a realistic transfer scenario where our analysis is computationally tractable, but bigger models trained on bigger datasets with more classes achieve better performance. These bigger datasets with richer label spaces could potentially help to mitigate the trade-off between pretraining accuracy and feature quality, and may also have class-level overlap with many downstream tasks, making greater class separation helpful rather than harmful.

Our results establish that a wide variety of losses that improve over vanilla softmax cross-entropy lead to greater class separation, but this observation does not immediately lead to a recipe for higher pretraining accuracy. The relationship between class separation and pretraining accuracy is non-monotonic: squared error achieves the highest class separation but does not significantly outperform vanilla softmax cross-entropy.

\cutsectionup
\vspace{-0.35em}
\section{Conclusion}
\vspace{-0.35em}
\cutsectiondown
\label{sec:conclusion}

In this study, we find that the properties of a loss that yields good performance on the pretraining task are different from the properties of a loss that learns good generic features. Training objectives that lead to better performance on the pretraining task learn more invariant representations with greater class separation. However, these same properties are detrimental when fixed features are transferred to other tasks. Moreover, because different loss functions produce different representations only in later network layers, which change substantially during fine-tuning, gains in pretraining accuracy do not lead to gains when models are fine-tuned on downstream tasks.

Our work suggests opportunities for improving fixed feature representations in deep neural networks. We see no inherent reason that features learned by softmax cross-entropy should be optimal for transfer, but previous work has optimized for pretraining accuracy, rather than transferable features. With the increasing importance of transfer learning in deep learning, we believe that future research into loss functions should explicitly target and evaluate performance under transfer settings.
\section*{Acknowledgements}
We thank Pieter-Jan Kindermans for comments on the manuscript, and Geoffrey Hinton and David Fleet for useful discussions.

\putbib[main]
\clearpage\newpage
\onecolumn
\counterwithin{figure}{section}
\counterwithin{table}{section}
\appendix
\part*{Appendix}
\startcontents[sections]
\printcontents[sections]{l}{1}{\setcounter{tocdepth}{2}}

\clearpage
\FloatBarrier
\section{Details of training and hyperparameter tuning}
\label{app:all_training_details}

\subsection{Training and tuning neural networks}
\label{app:training_details}

We trained ImageNet models (ResNet-50~\citep{he2016deep,resnetv15torchblogpost,goyal2017accurate} ``v1.5''\footnote{The \texttt{torchvision} ResNet-50 model and the ``official'' TensorFlow ResNet both implement this architecture, which was first proposed by \citet{resnetv15torchblogpost} and differs from the ResNet v1 described by \citet{he2016deep} in performing strided convolution in the first $3\times 3$ convolution in each stage rather than the first $1 \times 1$ convolution. Our implementation initializes the $\gamma$ parameters of the last batch normalization layer in each block to 0, as in~\citet{goyal2017accurate}.}) %
with SGD with Nesterov momentum of 0.9 and a batch size 4096 and weight decay of $8\times 10^{-5}$ (applied to the weights but not batch norm parameters). After 10 epochs of linear warmup to a maximum learning rate of 1.6, we decayed the learning rate by a factor of 0.975 per epoch. We took an exponential moving average of the weights over training as in~\textcite{szegedy2016rethinking}, with a momentum factor of 0.9999. We used standard data augmentation comprising random crops of 10-100\% of the image with aspect ratios of 0.75 to 1.33 and random horizontal flips. At test time, we resized images to 256 pixels on their shortest side and took a $224 \times 224$ center crop. Each training run took approximately 1.5 hours on a 128-core TPU v2 node. Overall, the experiments in the main text reflect 72 total training runs, plus an approximately equal number of training runs used to tune hyperparameters.

To tune hyperparameters, we held out a validation set of 50,046 ImageNet training examples. We initially performed a set of training runs with a wide range of different parameters, and then narrowed the hyperparameter range to the range shown in Table~\ref{tab:imagenet_hyperparameters}. To further tune the hyperparameters and the epoch for early stopping, we performed 3 training runs per configuration.\footnote{Due to the large number of hyperparameter configurations, for squared error, we performed only 1 run per configuration to select hyperparameters, but 3 to select the epoch at which to stop. We manually narrowed the hyperparameter search range until all trained networks achieved similar accuracy. The resulting hyperparaameters performed better than those suggested by \citet{hui2020evaluation}.} After determining the hyperparameters, we trained models on the full training set. We note that early stopping is important to achieve maximal performance with our learning rate schedule, but does not affect the conclusions we draw regarding transferability and class separation, as we confirm in Appendices~\ref{app:training_dynamics_transfer} and \ref{app:training_dynamics}. %

\begin{table}[h]
  \caption{\textbf{Hyperparameters for ResNet-50 on ImageNet.}}
  \aftertablecaption
  \centering
  \footnotesize
  \begin{tabular}{lll}
    \toprule
    Loss/regularizer & Hyperparameters & Epochs\\
    \midrule
    Softmax & N/A  &     146\\
    Label smoothing & $\alpha = \{0.08, 0.09, \bm{0.1}, 0.11. 0.12\}$  & 180\\
    Sigmoid  & N/A  &  166\\
    Extra final layer $L^2$  & $\lambda_\text{final} = \{4\text{e-}4, 6\text{e-}4, \bm{8\text{e-}4}, 1\text{e-}3\}$  & 168 \\
    Dropout  & $\rho = \{0.6, 0.65, \bm{0.7}, 0.75, 0.8, 0.85\}$ & 172\\
    Logit penalty  & $\beta = \{5\text{e-}5, 1\text{e-}4, 2\text{e-}4, 4\text{e-}4, \bm{6\text{e-}4}, 8\text{e-}4\}$   &  180\\
    Logit normalization  & $\tau = \{0.03, \bm{0.04}, 0.05, 0.06\}$    & 152\\
    Cosine softmax  & $\tau = \{0.04, 0.045, \bm{0.05}, 0.06, 0.07, 0.08\}$ & 158\\
    Squared error  &
    $\kappa = 9$, $M = 60$, $\text{loss scale} = 10$ & 196\\
    \bottomrule
  \end{tabular}
  \label{tab:imagenet_hyperparameters}
\end{table}

\subsection{Training and tuning multinomial logistic regression classifiers}
\label{app:logistic_training}
To train multinomial logistic regression classifiers on fixed features, we use L-BFGS~\cite{nocedal1980updating}, following a similar approach to previous work~\cite{kornblith2019better,radford2021learning}. We first extracted features for every image in the training set, by resizing them to 224 pixels on the shortest side and taking a $224\times 224$ pixel center crop. We held out a validation set from the training set, and used this validation set to select the $L^2$ regularization hyperparameter, which we selected from 45 logarithmically spaced values between $10^{-6}$ and $10^5$, applied to the sum of the per-example losses. Because the optimization problem is convex, we used the previous weights as a warm start as we increased the $L^2$ regularization hyperparameter. After finding the optimal hyperparameter on this validation set, we retrained on the training + validation sets and evaluated accuracy on the test set. We measured either top-1 or mean per-class accuracy, depending on which was suggested by the dataset creators. See Table~\ref{tab:dataset} for further details of the datasets investigated.

\begin{table}[h]
    \caption{\textbf{Datasets examined in transfer learning.}}
    \label{tab:dataset}
    \label{datasets}
    \begin{minipage}{\linewidth}
        \centering
        \footnotesize
        \begin{tabular}{llrrrrl}
        \toprule
        Dataset  &  \multicolumn{1}{l}{Classes} & \multicolumn{1}{l}{Size (train/test)} & Accuracy measure\\
        \midrule
        Food-101 \cite{bossard2014food} & 101 & 75,750/25,250 & top-1\\
        CIFAR-10 \cite{krizhevsky2009learning} & 10 & 50,000/10,000 & top-1\\
        CIFAR-100 \cite{krizhevsky2009learning}& 10 & 50,000/10,000 & top-1\\
        Birdsnap \cite{berg2014birdsnap} & 500 & 47,386/2,443 & top-1\\
        SUN397 \cite{xiao2010sun} & 397 & 19,850/19,850 & top-1\\
        Stanford Cars \cite{krause2013collecting} & 196 & 8,144/8,041 & top-1\\
        Oxford-IIIT Pets \cite{parkhi2012cats} & 37 & 3,680/3,369 & mean per-class\\
        Oxford 102 Flowers \cite{nilsback2008automated} & 102 & 2,040/6,149 & mean per-class\\
        \bottomrule
        \end{tabular}
    \end{minipage}
\end{table}

\subsection{Fine-tuning}
\label{app:fine_tuning}
In our fine-tuning experiments in Table~\ref{tab:finetune}, we used standard ImageNet-style data augmentation and trained for 20,000 steps with SGD with momentum of 0.9 and cosine annealing~\citep{loshchilov2016sgdr} without restarts. We performed hyperparameter tuning on a validation set, selecting learning rate values from a logarithmically spaced grid of 8 values between $10^{-5.5}$ and $10^{-1}$ and weight decay values from a logarithmically spaced grid of 8 values between $10^{-6.5}$ and $10^{-3}$, as well as no weight decay, dividing the weight decay by the learning rate. We manually verified that optimal hyperparameter combinations for each loss and dataset fall inside this grid. We averaged the accuracies obtained by hyperparameter tuning over 3 runs starting from 3 different pretrained ImageNet models and picked the best. We then retrained each model on combined training + validation sets and tested on the provided test sets.

\section{Confirmation of main findings with Inception v3}
\label{app:inception}

To confirm that our findings hold across architectures, we performed experiments using Inception v3~\cite{szegedy2016rethinking}, which does not have residual connections but still attains good performance on ImageNet ILSVRC. Because our goal was to validate the consistency of our observations, rather than to achieve maximum accuracy, we used the same hyperparameters as for ResNet-50, but selected the epoch for early stopping on a holdout set.

\begin{figure*}[t!]
\begin{center}
      \includegraphics[width=\textwidth]{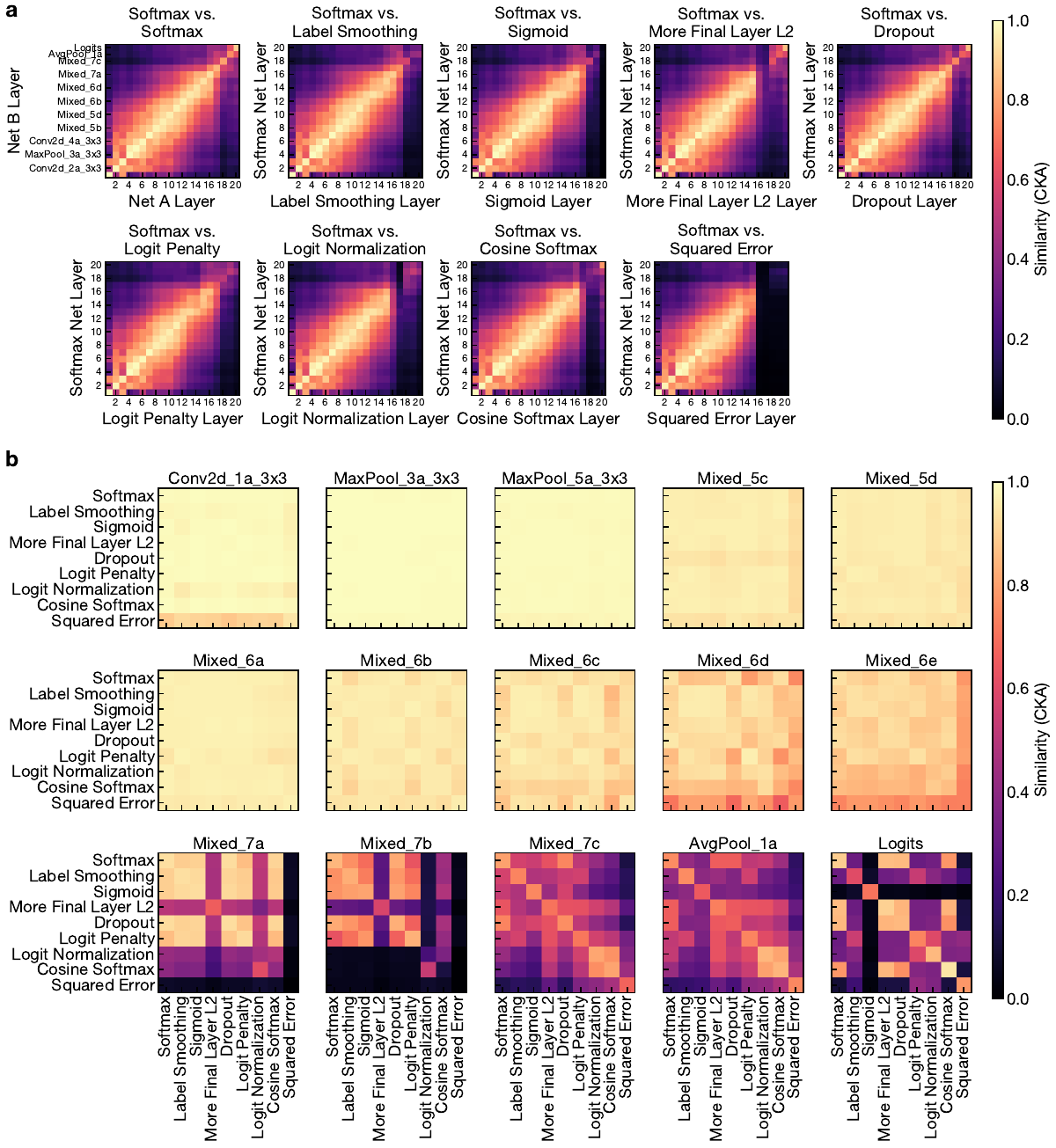}
\end{center}
\vskip -0.6em
\caption{\textbf{The choice of loss function affects representations only in later network layers, for Inception v3.} All plots show linear centered kernel alignment (CKA) between representations computed on the ImageNet validation set. \textbf{a}: CKA between network layers, for pairs of Inception v3 models trained from different initializations with the same or different losses. \textbf{b}: CKA between representations extracted from corresponding layers of networks trained with different loss functions. Diagonal reflects similarity of networks with the same loss function trained from different initalizations. See Figure~\ref{fig:cka} for results with ResNet-50.}
\label{fig:cka_inception}
\vskip -0.5em
\end{figure*}

Table~\ref{tab:transfer_inception} confirms our main findings involving class separation and transfer accuracy. As in Table~\ref{tab:transfer}, we observe that softmax learns more transferable features than other loss functions, and as in Table~\ref{tab:r2}, we find that lower class separation is associated with greater transferability. Figure~\ref{fig:cka_inception} confirms our finding that the choice of loss function affects representations only in later layers of the network.

\begin{table*}[h]
  \caption{\textbf{Objectives that produce higher ImageNet accuracy lead to less transferable fixed features, for Inception v3.}  ``ImageNet'' columns reflect accuracy of each model on the ImageNet validation set. ``Transfer'' columns reflect accuracy of $L^2$-regularized multinomial logistic regression classifiers trained to classify different transfer datasets using the fixed penultimate layer features of the ImageNet-trained networks. Numbers are averaged over 3 different pretraining initializations, and all values not significantly different than the best are bold-faced ($p < 0.05$, $t$-test). The strength of $L^2$ regularization is selected on a validation set that is independent of the test set. %
  See Table~\ref{tab:transfer} for results with ResNet-50.}
  \aftertablecaption
  \centering
  \fontsize{8}{10}\selectfont
  \setlength{\tabcolsep}{0.4em}
\begin{tabular}{lrrrrrrrrrrr}
\toprule
\multicolumn{1}{c}{} & \multicolumn{3}{c}{ImageNet}  & \multicolumn{8}{c}{Transfer}
\\
\cmidrule(lr){2-4}
\cmidrule(lr){5-12}
\multicolumn{1}{c}{Loss} & \multicolumn{1}{c}{Top-1} & \multicolumn{1}{c}{Top-5} & \multicolumn{1}{c}{$R^2$} &         Food &         CIFAR10 &        CIFAR100 &        Birdsnap &          SUN397 &   Cars &      Pets & \multicolumn{1}{c}{Flowers}\\
\midrule
Softmax             & 78.6 & 94.24 & 0.356 &  \textbf{74.5} &  \textbf{92.4} &  \textbf{76.2} &  \textbf{59.3} &  \textbf{63.1} &  \textbf{64.4} &           92.2 &  \textbf{94.0} \\
Label smoothing     & 78.8 & \textbf{94.60} & 0.441 &          73.3 &           91.6 &  \textbf{75.0} &           56.1 &           62.4 &           60.3 &  \textbf{93.0} &           92.4 \\
Sigmoid             & \textbf{79.1} & 94.17 & 0.444 &          73.7 &           91.3 &           74.7 &           55.0 &           62.0 &           60.7 &  \textbf{92.8} &  \textbf{93.0} \\
More final layer $L^2$ & \textbf{79.0} & 94.52 & 0.586 &         70.1 &           91.0 &           73.3 &           52.4 &           61.0 &           51.1 &  \textbf{92.5} &           89.6 \\
Dropout             & \textbf{79.0} & \textbf{94.50} & 0.454 &           72.6 &           91.5 &           74.7 &           56.3 &           62.1 &           59.2 &  \textbf{92.7} &           92.2 \\
Logit penalty       & \textbf{78.9} & \textbf{94.63} & 0.638 &           69.1 &           90.6 &           72.1 &           49.3 &           59.2 &           52.3 &           92.3 &           87.9 \\
Logit normalization & \textbf{78.8} & 94.34 & 0.559 &           67.4 &           90.6 &           72.2 &           50.9 &           58.5 &           45.6 &           92.1 &           84.2 \\
Cosine softmax      & \textbf{78.9} & 94.38 & 0.666 &           63.1 &           90.3 &           71.5 &           45.8 &           55.6 &           38.0 &           90.6 &           75.2 \\
Squared error       & 77.7 & 93.28 & 0.838 &           45.3 &           84.1 &           57.6 &           25.0 &           41.1 &           18.8 &           85.7 &           54.8 \\
\bottomrule
\end{tabular}
  \label{tab:transfer_inception}
\end{table*}

\FloatBarrier
\clearpage
\section{Additional evaluation of regularizers and losses}
\label{app:other_models}

\subsection{Training accuracy and learning curves}
\begin{table}[h!]
    \centering
  \caption{\textbf{Training accuracy of ResNet-50 models.} Regularizers and modified losses resulted in lower ImageNet training set accuracy, consistent with the notion that regularization sacrifices training accuracy to attain greater test accuracy. However, label smoothing was statstically tied with vanilla softmax cross-entropy in terms of training top-1 accuracy, and performed slightly better in terms of training top-5 accuracy. See Table~\ref{tab:transfer} for validation set accuracy.}
  \aftertablecaption
  \begin{tabular}{lll}
    \toprule
    Loss/regularizer & Top-1 Acc. (\%)     & Top-5 Acc. (\%) \\
    \midrule
    Softmax & $93.61 \pm 0.01$ & $99.33 \pm 0.002$ \\
    Label smoothing & $93.62 \pm 0.04$ & $99.43 \pm 0.007$\\
    Sigmoid  & $93.22 \pm 0.01$ & $99.19 \pm 0.002$  \\
    Extra final layer $L^2$  & $91.62 \pm 0.01$ & $98.85 \pm 0.003$  \\
    Dropout  & $92.25 \pm 0.01$ & $99.03 \pm 0.003$  \\
    Logit penalty  & $93.04 \pm 0.01$ & $99.13 \pm 0.002$  \\
    Logit normalization  & $92.86 \pm 0.01$ & $99.01 \pm 0.003$  \\
    Cosine softmax  & $92.47 \pm 0.01$ & $98.75 \pm 0.004$  \\
    Squared error  & $91.65 \pm 0.01$ & $98.59 \pm 0.002$  \\
    \bottomrule
  \end{tabular}
  \label{tab:resnet50_training}
\end{table}

\begin{figure}[h!]
\begin{center}
      \includegraphics[width=\textwidth]{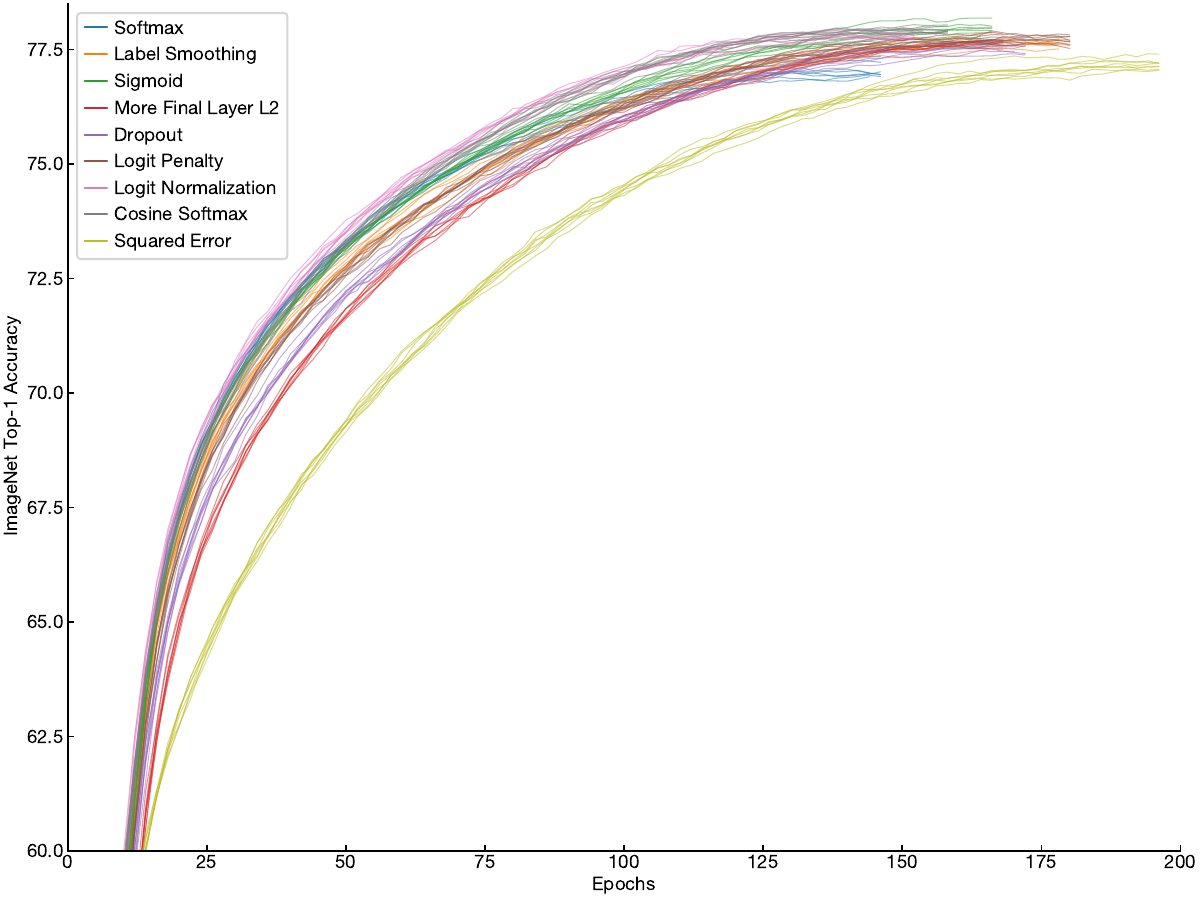}
\end{center}
\vskip -1em
\caption{\textbf{Evolution of ImageNet validation accuracy over training.} Each curve represents a different model. For each loss function, curves terminate at the epoch that provided the highest holdout set accuracy. Validation accuracy rises rapidly due to the use of an exponential moving average of the weights for evaluation. Some loss functions, such as logit normalization, appear to provide higher accuracy than vanilla softmax cross-entropy over the entire training run.}
\label{fig:validation_accuracy_over_training}
\end{figure}

\begin{figure}[h!]
\vskip -1em
\begin{center}
      \includegraphics[width=\textwidth]{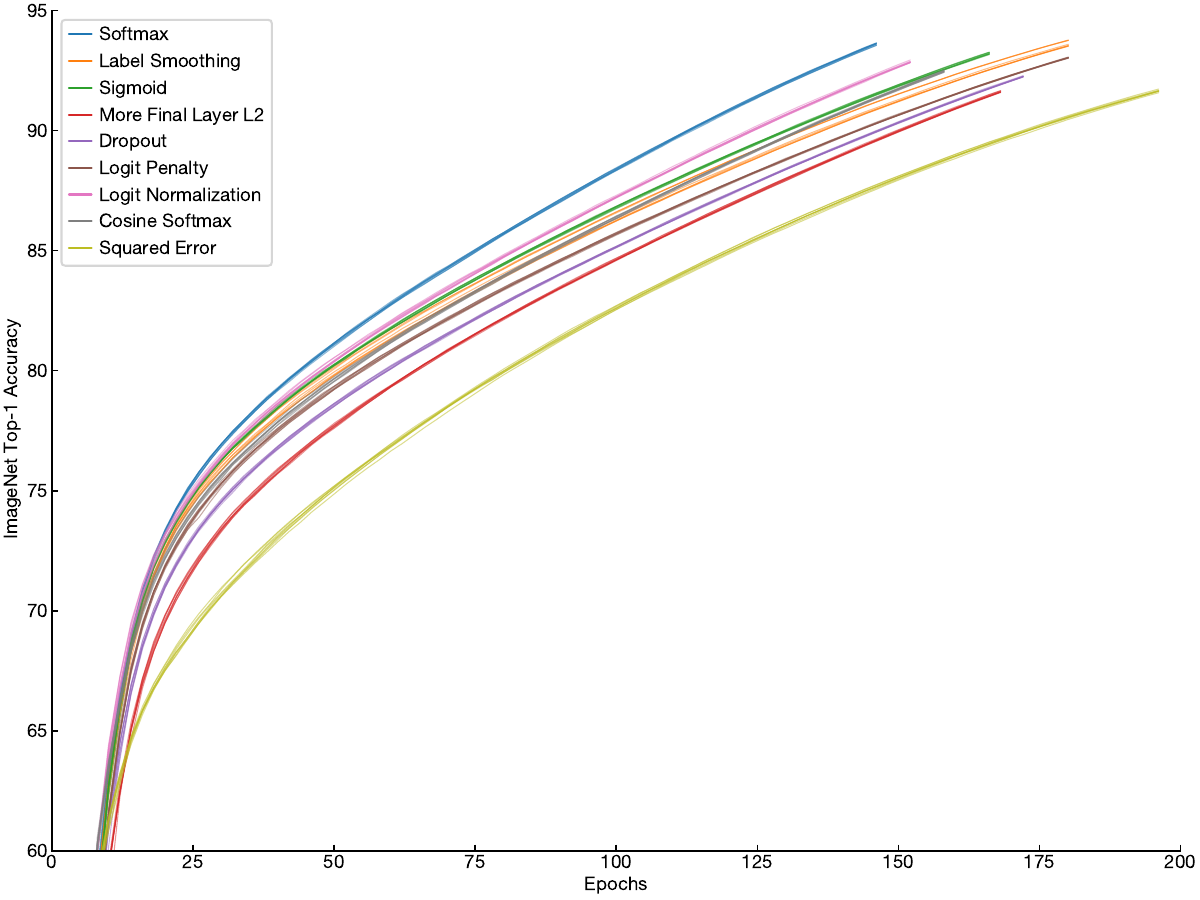}
\end{center}
\vskip -1em
\caption{\textbf{Evolution of ImageNet training accuracy.} Each curve represents a different model. For each loss function, curves terminate at the epoch that provided the highest holdout set accuracy.}
\label{fig:training_accuracy_over_training}
\end{figure}

\begin{figure}[h!]
\begin{center}
      \includegraphics[width=\textwidth]{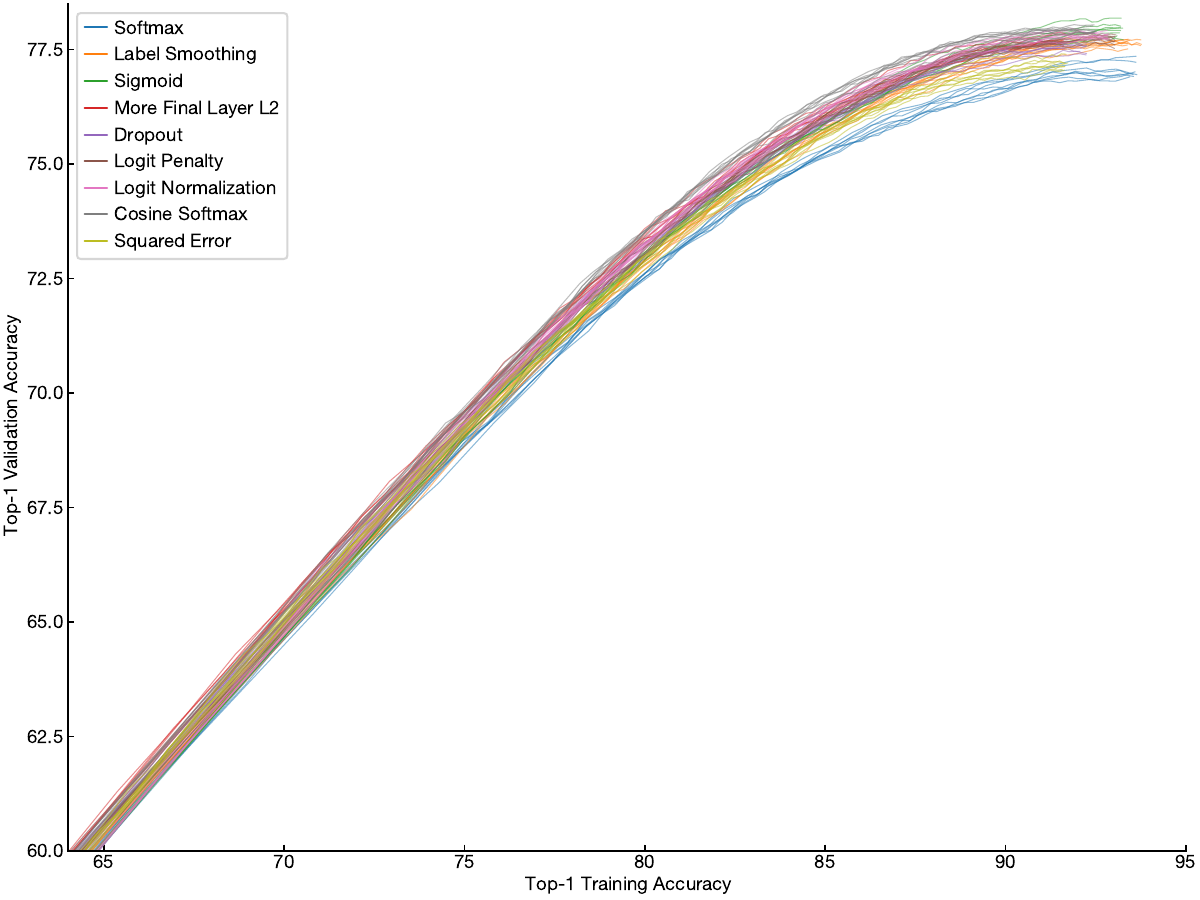}
\end{center}
\vskip -1em
\caption{\textbf{Validation versus training accuracy.} Each curve represents a different model. For each loss function, curves terminate at the training accuracy that provided the highest holdout set accuracy. Regularized models achieve higher validation accuracy at a given training accuracy as compared to softmax.}
\label{fig:training_accuracy_vs_val}
\end{figure}

\FloatBarrier
\subsection{Robustness and calibration}

In addition to the differences in class separation and accuracy described in the text, losses differed in out-of-distribution robustness, and in the calibration of the resulting predictions. Table~\ref{tab:ood} shows results for ImageNet-trained ResNet-50 models on the out-of-distribution test sets ImageNet-V2~\citep{recht2019imagenet}, ImageNet-A~\citep{hendrycks2019natural}, ImageNet-Sketch~\citep{wang2019learning}, ImageNet-R~\citep{hendrycks2020many}, and ImageNet-C~\citep{hendrycks2019benchmarking}. In almost all cases, alternative loss functions outperformed softmax cross-entropy, with logit normalization and cosine softmax typically performing slightly better than alternatives. Effects on calibration, shown in Table~\ref{tab:calibration}, were mixed. Label smoothing substantially reduced expected calibration error~\citep{guo2017calibration}, as previously shown by~\citet{muller2019does}, although cosine softmax achieved a lower negative log likelihood. However, there was no clear relationship between calibration and accuracy. Logit penalty performed well in terms of accuracy, but provided the worst calibration of any objective investigated.
\begin{table}[h]
    \caption{\textbf{Regularizers and alternative losses improve performance on out-of-distribution test sets.} Accuracy averaged over 8 ResNet-50 models per loss.}
  \aftertablecaption
  \setlength{\tabcolsep}{0pt}
  \centering
  \scriptsize
  \begin{tabular}{lrrrrr}
    \toprule
    Loss/regularizer & \multicolumn{1}{p{2.3cm}}{\hfill ImageNet-V2 (\%)} &
    \multicolumn{1}{p{2.4cm}}{\hfill ImageNet-A (\%)} & \multicolumn{1}{p{2.4cm}}{\hfill IN-Sketch (\%)} & \multicolumn{1}{p{2.4cm}}{\hfill ImageNet-R (\%)} & \multicolumn{1}{p{2.4cm}}{\hfill ImageNet-C (mCE)}\\
    \midrule
Softmax
& $65.0 \pm 0.1$
& $2.7 \pm 0.0$
& $21.8 \pm 0.1$
& $36.8 \pm 0.1$
& $75.9 \pm 0.1$
\\
Label smoothing
& $\mathbf{65.7} \pm 0.1$
& $3.8 \pm 0.1$
& $22.5 \pm 0.1$
& $37.8 \pm 0.1$
& $75.2 \pm 0.1$
\\
Sigmoid
& $\mathbf{65.9} \pm 0.1$
& $3.3 \pm 0.0$
& $22.6 \pm 0.1$
& $36.6 \pm 0.1$
& $74.6 \pm 0.1$
\\
Extra final layer $L^2$
& $\mathbf{65.8} \pm 0.1$
& $3.3 \pm 0.0$
& $23.1 \pm 0.1$
& $37.7 \pm 0.1$
& $74.1 \pm 0.1$
\\
Dropout
& $65.4 \pm 0.0$
& $3.1 \pm 0.1$
& $23.0 \pm 0.1$
& $37.2 \pm 0.1$
& $74.5 \pm 0.1$
\\
Logit penalty
& $\mathbf{65.8} \pm 0.0$
& $4.5 \pm 0.0$
& $22.8 \pm 0.1$
& $38.1 \pm 0.1$
& $74.3 \pm 0.1$
\\
Logit normalization
& $\mathbf{65.8} \pm 0.1$
& $\mathbf{4.8} \pm 0.1$
& $23.7 \pm 0.1$
& $\mathbf{39.2} \pm 0.1$
& $73.2 \pm 0.1$
\\
Cosine softmax
& $\mathbf{65.8} \pm 0.1$
& $\mathbf{4.6} \pm 0.1$
& $\mathbf{24.8} \pm 0.1$
& $38.7 \pm 0.1$
& $\mathbf{72.5} \pm 0.1$
\\
Squared error
& $65.3 \pm 0.1$
& $4.5 \pm 0.1$
& $22.4 \pm 0.1$
& $36.3 \pm 0.1$
& $74.6 \pm 0.1$
\\
    \bottomrule
  \end{tabular}
  \label{tab:ood}
\end{table}

\begin{table}[h]
    \caption{\textbf{Some regularizers and alternative losses improve calibration.} We report negative log likelihood (NLL) and expected calibration error (ECE), averaged over 3 ResNet-50 models trained with each loss on the ImageNet validation set, before and after scaling the temperature of the probability of the distribution to minimize NLL, as in~\citet{guo2017calibration}. These models were trained with a holdout set of 50,046 ImageNet training examples, which were then used to perform temperature scaling to minimize NLL. ECE is computed with 15 evenly spaced bins. For networks trained with sigmoid loss, we normalize the probability distribution by summing probabilities over all classes.}
  \aftertablecaption
  \centering
  \begin{tabular}{lrrrr}
    \toprule
    & \multicolumn{2}{c}{Uncalibrated} & \multicolumn{2}{c}{With temperature scaling}\\
    \cmidrule(lr){2-3} \cmidrule(lr){4-5}
    Loss/regularizer & \multicolumn{1}{c}{NLL} & \multicolumn{1}{c}{ECE} & \multicolumn{1}{c}{NLL} & \multicolumn{1}{c}{ECE} \\
    \midrule
    Softmax & $0.981 \pm 0.002$ & $0.073 \pm 0.0001$ & $0.917 \pm 0.002$ & $0.027 \pm 0.0004$\\
    Label smoothing & $0.947 \pm 0.001$ & $\mathbf{0.016} \pm 0.0007$ & $0.941 \pm 0.001$ & $0.044 \pm 0.0004$\\
    Sigmoid  & $0.944 \pm 0.002$ & $0.044 \pm 0.0003$ & $0.914 \pm 0.002$ & $\mathbf{0.019} \pm 0.0002$ \\
    Extra final layer $L^2$  & $0.976 \pm 0.002$ & $0.081 \pm 0.0003$ & $0.908 \pm 0.002$ & $0.038 \pm 0.0006$ \\
    Dropout  & $0.971 \pm 0.002$  & $0.074 \pm 0.0009$  & $0.905 \pm 0.002$ & $0.031 \pm 0.0002$ \\
    Logit penalty  & $1.041 \pm 0.001$ & $0.090 \pm 0.0003$ & $0.995 \pm 0.001$ & $0.055 \pm 0.0004$ \\
    Logit normalization  & $0.965 \pm 0.001$ & $0.069 \pm 0.0002$ & $0.949 \pm 0.001$ & $0.049 \pm 0.0003$ \\
    Cosine softmax  & $\mathbf{0.912} \pm 0.002$ & $0.066 \pm 0.0006$  & $\mathbf{0.895} \pm 0.002$ & $0.043 \pm 0.0008$\\
    \bottomrule
  \end{tabular}
  \label{tab:calibration}
\end{table}

\FloatBarrier
\subsection{Similarity of model predictions}

\begin{figure*}[h!]
\begin{center}
      \includegraphics[width=\textwidth]{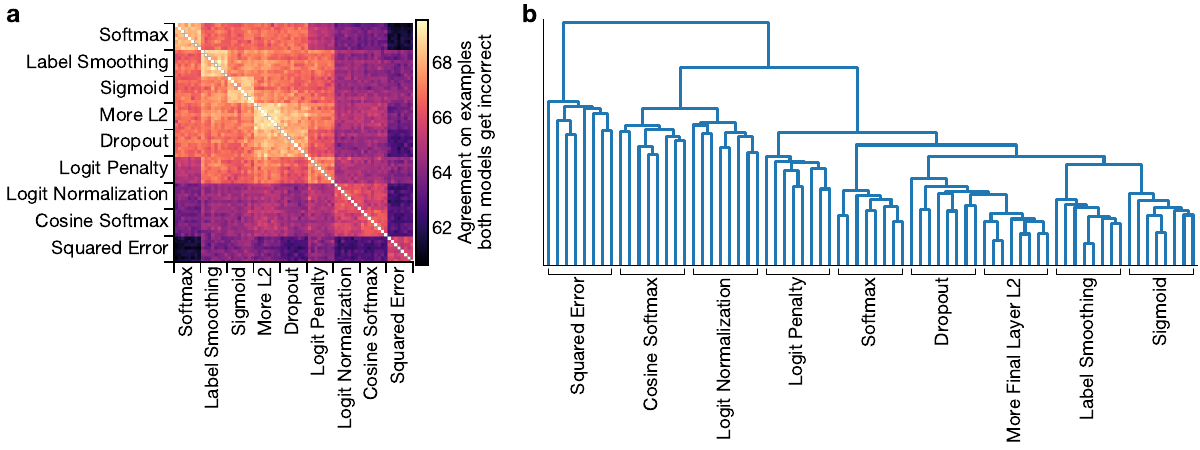}
\end{center}
\vskip -1.0em
\caption{\textbf{Different losses produce different predictions, even when accuracies are close.} \textbf{a}: For each pair of models, we take examples incorrectly classified by both and measure the percentage where the models' top-1 predictions agree. We show results for 8 different initializations trained with each objective. See Figure~\ref{fig:prediction_similarity_additional_metrics} for qualitatively similar plots that show percentages of all examples on which models agree, and percentages of images where both models are either correct or incorrect. \textbf{b}: Dendrogram based on similarity matrix. All models naturally cluster according to loss, except for ``Dropout'' and ``More Final Layer L2'' models.}
\label{fig:prediction_similarity}
\end{figure*}

\begin{figure}[h!]
\begin{center}
      \includegraphics[width=0.9\textwidth]{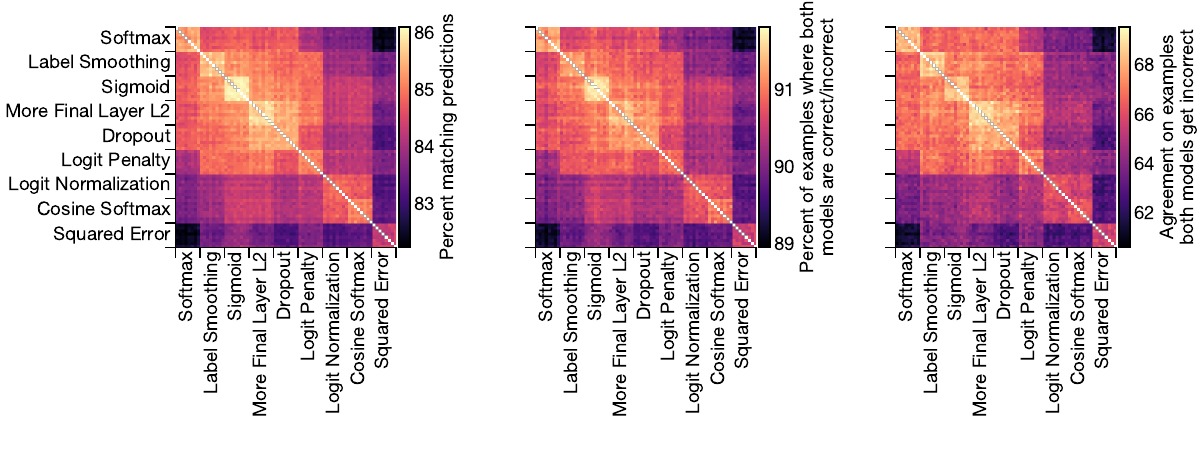}
\end{center}
\vskip -1.5em
\caption{\textbf{Different ways of measuring similarity of single-model ResNet-50 predictions yield similar qualitative results.} In the left panel, we compute the top-1 predictions for pairs of models on the ImageNet validation set and determine the percentage of examples where these predictions match. In the middle panel, we measure the percentage of examples where models either get both right or both wrong. In the right panel, we restrict our analysis to examples that both models get incorrect, and measure the percentage of these examples where both models make the same (incorrect) top-1 prediction.}
\label{fig:prediction_similarity_additional_metrics}
\vskip -1em
\end{figure}

Given that many loss functions resulted in similar improvements in accuracy over softmax loss, we sought to determine whether they also produced similar effects on network predictions. For each pair of models, we selected  validation set examples that both models classified incorrectly, and measured the percentage of these examples for which the models gave the same prediction. As shown in Figure~\ref{fig:prediction_similarity}, models' predictions cluster into distinct groups according to their objectives. Models trained with the same objective (from different initializations) are more similar than models trained with different objectives. In addition, models trained with (regularized) softmax loss or sigmoid loss are more similar to each other than to models trained with logit normalization or cosine softmax, and networks trained with squared error are dissimilar to all others examined. Figure~\ref{fig:prediction_similarity_additional_metrics} shows that other ways of measuring the similarity of models' predictions yielded qualitatively similar results.

\label{app:model_predictions}

\begin{figure}[h]
\begin{center}
      \includegraphics[width=0.9\textwidth]{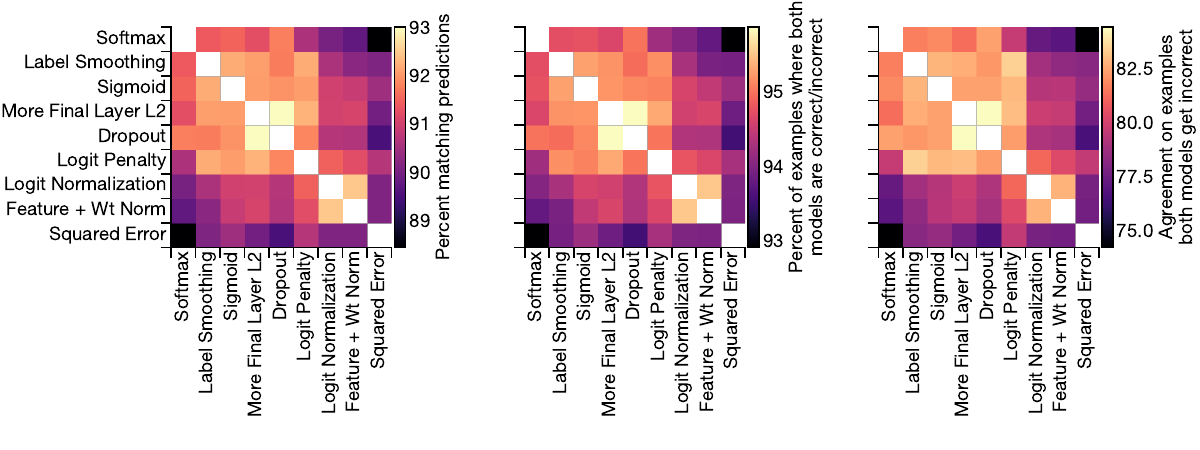}
\end{center}
\vskip -1.5em
\caption{\textbf{Ensemble predictions are substantially more similar than single-model predictions.} Predictions of the ensemble were computed by taking 8 ResNet-50 models trained from different random initializations with the same loss and picking the most common top-1 prediction for each example.}
\label{fig:ensemble_similarity_metrics}
\end{figure}

\begin{figure}[h]
\begin{center}
      \includegraphics[width=0.45\textwidth]{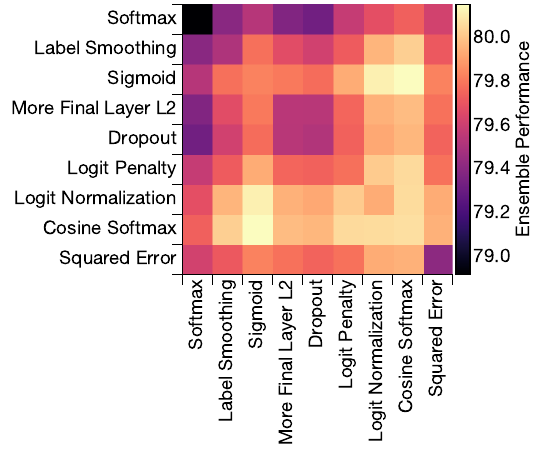}
\end{center}
\vskip -1.0em
\caption{\textbf{Ensembling models trained with different losses provides only modest performance benefits.} Ensembles consist of 8 ResNet-50 models, half of which are trained with the objective on the x-axis, the other half with the objective on the y-axis. The ensemble prediction is the modal class prediction of the 8 models.}
\label{fig:ensembling}
\vskip -1em
\end{figure}

Although it was possible to identify the loss used to train individual models from their predictions, models trained with the same loss nonetheless disagreed on many examples. Standard deviations in top-1 accuracy are $<$0.2\% for all losses, but even the most similar pair of models provides different predictions on 13.9\% of all validation set examples (Figure~\ref{fig:prediction_similarity_additional_metrics}). Ensembling can substantially reduce the level of disagreement between models and objectives: When ensembling the 8 models trained with the same loss, the least similar losses (softmax and squared error) disagree on only 11.5\% of examples (Figure~\ref{fig:ensemble_similarity_metrics}). However, there was little accuracy benefit to ensembling models trained with different objectives over ensembling models trained with the same objective (Figure~\ref{fig:ensembling}). %

\FloatBarrier
\subsection{Combining regularizers does not improve accuracy}
\label{app:combined_losses}

Given the clear differences in the effects of different objectives on network predictions, we next asked whether combining regularization or normalization strategies might result in better predictions. Table~\ref{tab:combined_losses} shows that these combinations do not improve accuracy.
However, as shown in Table~\ref{tab:autoaugment}, improved data augmentation~\cite{cubuk2019autoaugment,zhang2017mixup} provides a similar additive gain in accuracy to networks trained with alternative losses as it does to networks trained with softmax cross-entropy.
These results suggest that the objectives that improve over softmax cross-entropy do so via similar mechanisms, but data augmentation acts differently. %

\begin{table}[h!]
  \caption{\textbf{Combining final-layer regularizers and/or improved losses does not meaningfully enhance performance.} Accuracy of ResNet-50 models on our ImageNet holdout set when combining losses and regularizers between models. All results reflect the maximum accuracy on the holdout set at any point during training, averaged across 3 training runs. Accuracy numbers are higher on the holdout set than the official ImageNet validation set. This difference in accuracy is likely due to a difference in image distributions between the ImageNet training and validation sets, as previously noted in Section C.3.1 of \textcite{recht2019imagenet}.}
  \aftertablecaption
  \centering
  \begin{tabular}{lrrrr}
    \toprule
    &\multicolumn{1}{c}{Baseline} &
     \multicolumn{1}{p{2.5cm}}{\centering Label smoothing ($\alpha = 0.1$)}
     & \multicolumn{1}{c}{Sigmoid} & \multicolumn{1}{p{2.5cm}}{\centering Cosine softmax ($\tau = 0.05$)} \\
    \midrule
    Baseline & $79.9$ & $80.4$ & $80.6$ & $80.6$\\
    Dropout ($\beta = 0.7$) & $80.3$ & $80.3$ & $80.3$ & $80.2$\\
    Dropout ($\beta = 0.8$) & $80.2$ & $80.4$ & $80.4$ & $80.4$\\
    Dropout ($\beta = 0.9$) &  & $80.3$ & $80.5$ & $80.6$\\
    Dropout ($\beta = 0.95$) &  & $80.4$ & $80.6$ & $80.7$\\
    Logit penalty ($\gamma = 5 \times 10^{-5}$) & $80.4$ & $80.3$ & $80.5$ & $80.6$\\
    Logit penalty ($\gamma = 1 \times 10^{-4}$) & $80.4$ & $80.3$ & $80.5$ & $80.5$\\
    Logit penalty ($\gamma = 2 \times 10^{-4}$) & $80.4$ & $80.3$ & $80.4$ & $80.5$\\
    Logit penalty ($\gamma = 4 \times 10^{-4}$) & $80.4$ & $80.2$ & $80.3$ & $80.5$\\
    Logit penalty ($\gamma = 6 \times 10^{-4}$) & $80.5$ & $80.2$ & $80.3$ & $80.5$\\
    Logit normalization ($\tau = 0.02$) & & $80.0$ & $80.4$ & \\
    Logit normalization ($\tau = 0.03$) & $80.3$ & $80.4$ & $80.6$ & \\
    Logit normalization ($\tau = 0.04$) & $80.4$ & $80.5$ & $80.6$ & \\
    Logit normalization ($\tau = 0.05$) & $80.3$ & $80.5$ & $80.5$ & \\
    Logit normalization ($\tau = 0.06$) & $80.3$ & $80.4$ & $80.5$ & \\
    Cosine normalization ($\tau = 0.045$) & $80.6$ & & $80.5$ & \\
    Cosine normalization ($\tau = 0.05$) & $80.6$ & & $80.6$ & \\
    Cosine normalization ($\tau = 0.06$) & $80.4$ & & $75.3$ & \\
    \bottomrule
  \end{tabular}
  \label{tab:combined_losses}
\end{table}

\begin{table}[h!]
  \caption{\textbf{AutoAugment and Mixup provide consistent accuracy gains beyond well-tuned losses and regularizers.} Top-1 accuracy of ResNet-50 models trained with and without AutoAugment, averaged over 3 (with AutoAugment) or 8 (without AutoAugment) runs. Models trained with AutoAugment use the loss hyperparameters chosen for models trained without AutoAugment, but the point at which to stop training was chosen independently on our holdout set. For models trained with Mixup, the mixing parameter $\alpha$ is chosen from $[0.1, 0.2, 0.3, 0.4]$ on the holdout set. Best results in each column, as well as results insignificantly different from the best ($p > 0.05$, t-test), are bold-faced.}
  \aftertablecaption
  \centering
  \footnotesize
  \setlength{\tabcolsep}{0.45em}
  \begin{tabular}{lrrrrrr}
    \toprule
    & \multicolumn{2}{c}{Standard augmentation} & \multicolumn{2}{c}{AutoAugment} & \multicolumn{2}{p{3.5cm}}{\centering Mixup}\\
    \cmidrule(lr){2-3} \cmidrule(lr){4-5}  \cmidrule(lr){6-7}
    Loss/regularizer & Top-1 (\%)     & Top-5 (\%)  & Top-1 (\%)     & Top-5  (\%)  & Top-1 (\%)     & Top-5  (\%)\\
    \midrule
    Softmax & $77.0 \pm 0.06$  & $93.40 \pm 0.02$  & $77.7 \pm 0.05$ & $93.74 \pm 0.05$ &  $78.0 \pm 0.05$ & $93.98 \pm 0.03$ \\
    Sigmoid  & $\mathbf{77.9} \pm 0.05$  & $93.50 \pm 0.02$ & $\mathbf{78.5} \pm 0.04$ & $93.82 \pm 0.02$ & $\mathbf{78.5} \pm 0.07$ & $93.94 \pm 0.04$ \\
    Logit penalty  & $77.7 \pm 0.02$   & $\mathbf{93.83} \pm 0.02$  & $\mathbf{78.3} \pm 0.05$ & $\mathbf{94.10} \pm 0.03$ & $78.0 \pm 0.05$ & $93.95 \pm 0.05$\\
    Cosine softmax  & $\mathbf{77.9} \pm 0.02$   & $\mathbf{93.86} \pm 0.01$  & $\mathbf{78.3} \pm 0.02$ & $\mathbf{94.12} \pm 0.04$ & $\mathbf{78.4} \pm 0.04$ & $\mathbf{94.14} \pm 0.02$\\
    \bottomrule
  \end{tabular}
  \label{tab:autoaugment}
\end{table}

With longer training, both sigmoid cross-entropy and cosine softmax achieve state-of-the-art accuracy among ResNet-50 networks trained with AutoAugment (Table~\ref{tab:sota}), matching or outperforming supervised contrastive learning~\cite{khosla2020supervised}. Combining cosine softmax loss, AutoAugment, and Mixup, we achieve 79.1\% top-1 accuracy and 94.5\% top-5 accuracy, which was, at the time this paper was first posted, the best reported $224\times 224$ pixel single-crop accuracy with an unmodified ResNet-50 architecture trained from scratch.

\begin{table}[h!]
    \caption{\textbf{Comparison with state-of-the-art.} All results are for ResNet-50 models trained with AutoAugment. Loss hyperparameters are the same as in Table~\ref{tab:autoaugment}, but the learning schedule decays exponentially at a rate of 0.985 per epoch, rather than 0.975 per epoch. This learning rate schedule takes approximately $2\times$ as many epochs before it reaches peak accuracy, and provides a $\sim$0.4\% improvement in top-1 accuracy across settings.}
    \aftertablecaption
    \centering
  \begin{tabular}{lrrrrrr}
    \toprule
    Loss &
    Epochs & Top-1 (\%) & Top-5 (\%)\\
    \midrule
    Softmax \citep{cubuk2019autoaugment} & 270 & $77.6$ & $93.8$\\
    Supervised contrastive \citep{khosla2020supervised} & 700 & $\mathbf{78.8}$ & $93.9$\\
    \midrule
    \textit{Ours:}\\
    Softmax & 306 & $77.9 \pm 0.02$ & $93.77 \pm 0.03$\\
    Sigmoid & 324 & $\mathbf{78.9} \pm 0.04$ & $93.96 \pm 0.06$\\
    Logit penalty & 346 & $\mathbf{78.6} \pm 0.07$ & $\mathbf{94.30} \pm 0.01$\\
    Cosine softmax & 308 & $\mathbf{78.7} \pm 0.04$ & $\mathbf{94.24} \pm 0.02$\\
    \hline \hline
    \textit{Ours (with Mixup):}\\
    Sigmoid & 384 & $79.1 \pm 0.06$ & $94.28 \pm 0.03$\\
    Cosine softmax & 348 & $79.1 \pm 0.09$ & $94.49 \pm 0.01$\\
    \bottomrule
  \end{tabular}
  \label{tab:sota}
\end{table}

\FloatBarrier
\section{Additional transfer learning results}
\subsection{Scatterplots of ImageNet vs. transfer accuracy by dataset}
\begin{figure}[h!]
\begin{center}
      \includegraphics[width=\textwidth]{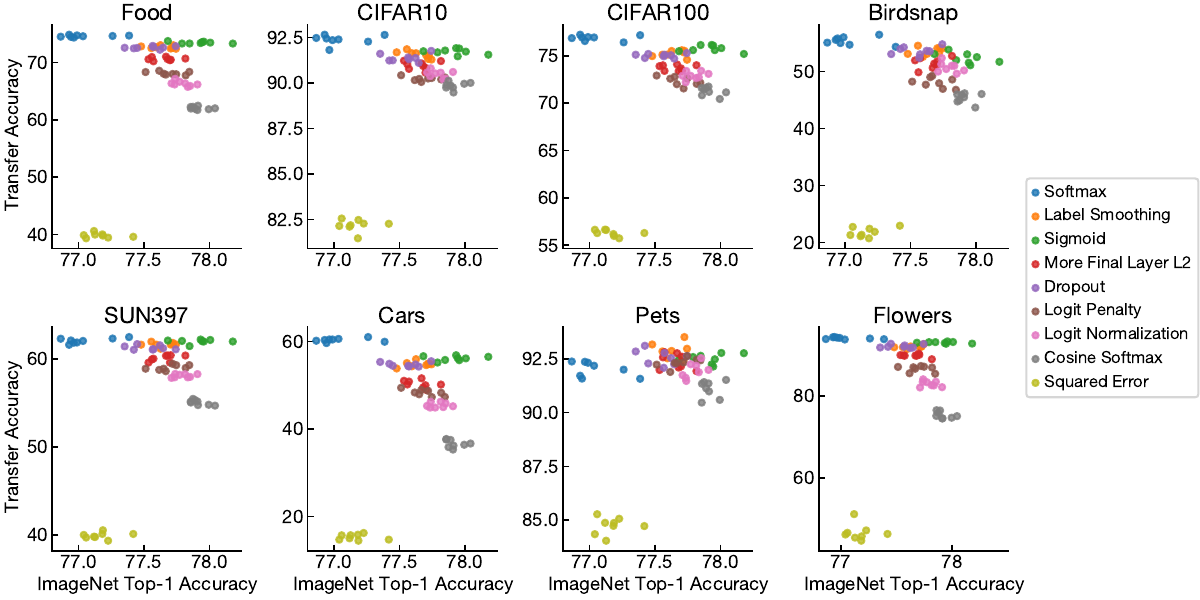}
\end{center}
\vskip -0.5em
\caption{\textbf{Higher ImageNet accuracy is not associated with higher linear transfer accuracy.} Points represent the accuracies of individual training runs. Panels represent different datasets. See Figure~\ref{fig:imagenet_vs_transfer} for a similar plot of transfer accuracy averaged across datasets.}
\label{fig:imagenet_vs_transfer_datasets}
\vskip -1em
\end{figure}

\subsection{Training dynamics of transfer accuracy}
\label{app:training_dynamics_transfer}
\begin{figure}[h!]
\begin{center}
      \includegraphics[width=\textwidth]{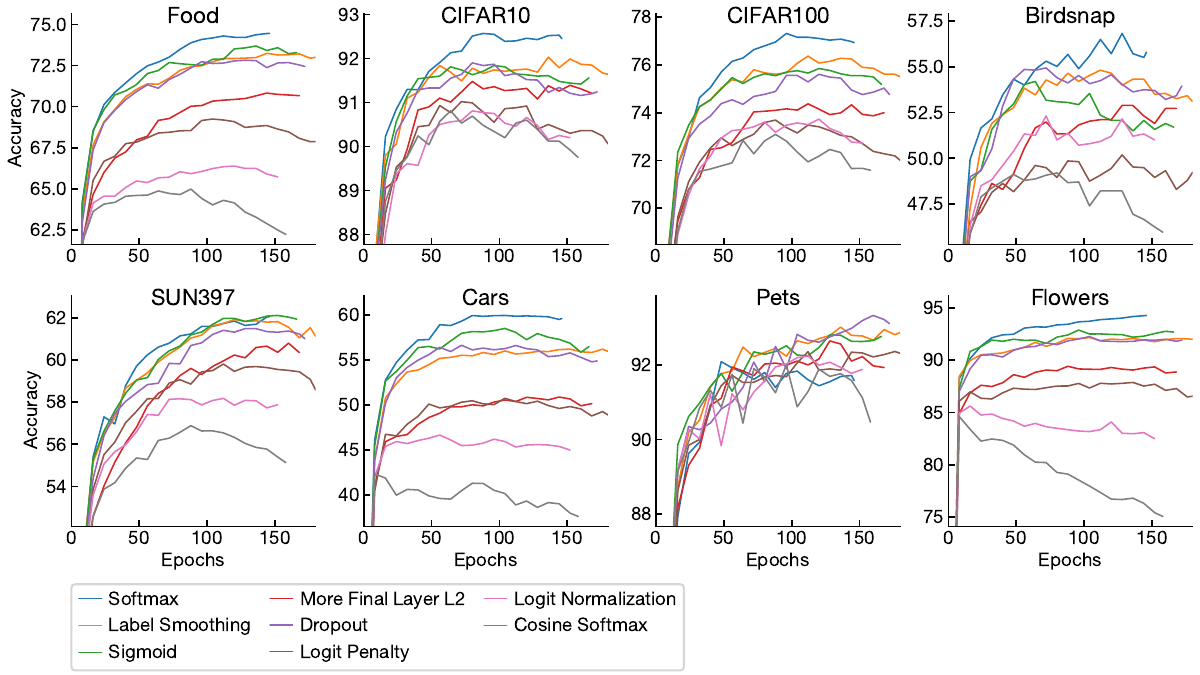}
\end{center}
\vskip -0.5em
\caption{\textbf{On most datasets, softmax cross-entropy achieves the highest linear transfer accuracy over the entire training run.} For each loss function, we evaluate the linear transfer accuracy of a ResNet-50 model every 6 epochs over the course of a single ImageNet pretraining run. Lines terminate at the final checkpoint, selected as described in Appendix~\ref{app:training_details}. Conclusions regarding the superiority of different loss functions match those from Table~\ref{tab:transfer}: Vanilla softmax cross-entropy achieves greater accuracy than other losses except on SUN397, where it is tied with sigmoid cross-entropy, and Pets, where other losses perform better. We do not show results for squared error because they are off the scale of the plots for all datasets except Pets.}
\label{fig:training_dynamics_transfer}
\vskip -1em
\end{figure}

\FloatBarrier
\subsection{Results on Chexpert}
\label{app:chexpert}
In Table~\ref{tab:chexpert}, we evaluate the performance of linear classifiers trained to classify the Chexpert chest X-ray dataset~\cite{irvin2019chexpert} based on the penultimate layer representations of our ImageNet-pretrained models, using the procedure described in Appendix~\ref{app:logistic_training}. We treat both uncertain and unmentioned observations as negative. We tune the $L^2$ regularization hyperparameter separately for each class. We approximate AUC using 1000 evenly-spaced bins. The official validation set of 234 images is very small and results in high variance; vanilla softmax cross-entropy achieves the best numerical results on all but one pathology, but many losses are statistically tied. We thus examine a second setting where we split 22,431 images from the training set and evaluate on these images. On this split, we find that softmax cross-entropy performs significantly better than all other losses on 4 of the 5 pathologies, and is tied for the best AUC on the fifth.

We note that the domain shift between Chexpert and ImageNet is very large. Given the extent of the domain shift, linear transfer will always perform far worse than fine-tuning. However, fine-tuning is unlikely to reveal differences among losses, particularly given that \citet{raghu2019transfusion} previously reported that ImageNet pretraining provides no accuracy advantage over training from scratch on this dataset. Nonetheless, we find that, even in this somewhat extreme setting, the fixed features learned by vanilla softmax cross-entropy on ImageNet work better than features learned by other losses.

\begin{table*}[h]
  \caption{\textbf{Transfer learning results on Chexpert.}  AUC of classifiers learned using $L^2$-regularized multinomial logistic regression on the fixed penultimate layer features of the ImageNet-trained networks. Numbers are averaged over 8 different pretraining initializations, and all values not significantly different than the best are bold-faced ($p < 0.05$, $t$-test). The strength of $L^2$ regularization is selected on a validation set that is independent of the test set. %
  See Table~\ref{tab:transfer} for results for natural image datasets.}
  \aftertablecaption
  \centering
  \setlength{\tabcolsep}{0.5em}
\begin{tabular}{lrrrrr}
\toprule
\multicolumn{1}{c}{Pretraining loss} & \multicolumn{1}{c}{Atelectasis} & \multicolumn{1}{c}{Cardiomegaly} & \multicolumn{1}{c}{Consolidation} & \multicolumn{1}{c}{Edema} & \multicolumn{1}{c}{Pleural effusion}\\
\midrule
\multicolumn{6}{l}{\textit{Official validation set (234 images):}}\\
Softmax         &  \textbf{74.9} &  \textbf{75.3} &  \textbf{86.9} &  \textbf{87.4} &  \textbf{86.4} \\
Label smoothing &  \textbf{74.0} &  \textbf{73.6} &  \textbf{85.8} &  \textbf{86.5} &  \textbf{85.8} \\
Sigmoid         &  \textbf{74.6} &  \textbf{75.0} &  \textbf{86.9} &  \textbf{87.3} &           85.3 \\
More final layer $L^2$        &  \textbf{74.6} &  \textbf{74.1} &  \textbf{85.2} &           85.2 &           84.8 \\
Dropout         &  \textbf{74.9} &  \textbf{73.3} &  \textbf{86.0} &  \textbf{86.3} &           84.0 \\
Logit penalty   &  \textbf{74.6} &  \textbf{75.9} &           83.5 &           84.8 &           83.3 \\
Logit normalization         &  \textbf{74.2} &  \textbf{73.6} &  \textbf{85.6} &           82.2 &           83.8 \\
Cosine softmax &  \textbf{73.3} &           71.9 &           83.5 &           81.6 &           82.2 \\
Squared error   &           71.2 &           67.5 &           76.2 &           73.0 &           75.0 \\
\midrule
\multicolumn{6}{l}{\textit{i.i.d. split from training set (22,431 images):}}\\
Softmax         &          \textbf{65.8} &          \textbf{76.2} &          \textbf{66.8} &          \textbf{79.9} &          \textbf{81.4} \\
Label smoothing &                   64.9 &                   74.9 &                   66.3 &                   79.2 &                   80.2 \\
Sigmoid         &                   65.4 &                   74.9 &          \textbf{66.5} &                   79.3 &                   80.5 \\
More final layer $L^2$        &                   64.7 &                   73.6 &                   65.8 &                   78.4 &                   79.7 \\
Dropout         &                   65.0 &                   74.6 &          \textbf{66.5} &                   79.0 &                   80.4 \\
Logit penalty   &                   64.1 &                   72.7 &                   65.3 &                   78.1 &                   78.9 \\
Logit normalization         &                   63.9 &                   72.0 &                   65.2 &                   77.6 &                   78.4 \\
Cosine softmax &                   62.6 &                   70.0 &                   64.2 &                   76.4 &                   76.7 \\
Squared error   &                   59.6 &                   64.9 &                   59.9 &                   72.2 &                   69.5 \\
\bottomrule
\end{tabular}
  \label{tab:chexpert}
\end{table*}

\clearpage
\FloatBarrier
\subsection{CKA between models before and after transfer}
\vskip -1em
\label{app:cka_transfer}
\begin{figure}[h!]
\begin{center}
      \includegraphics[width=\textwidth]{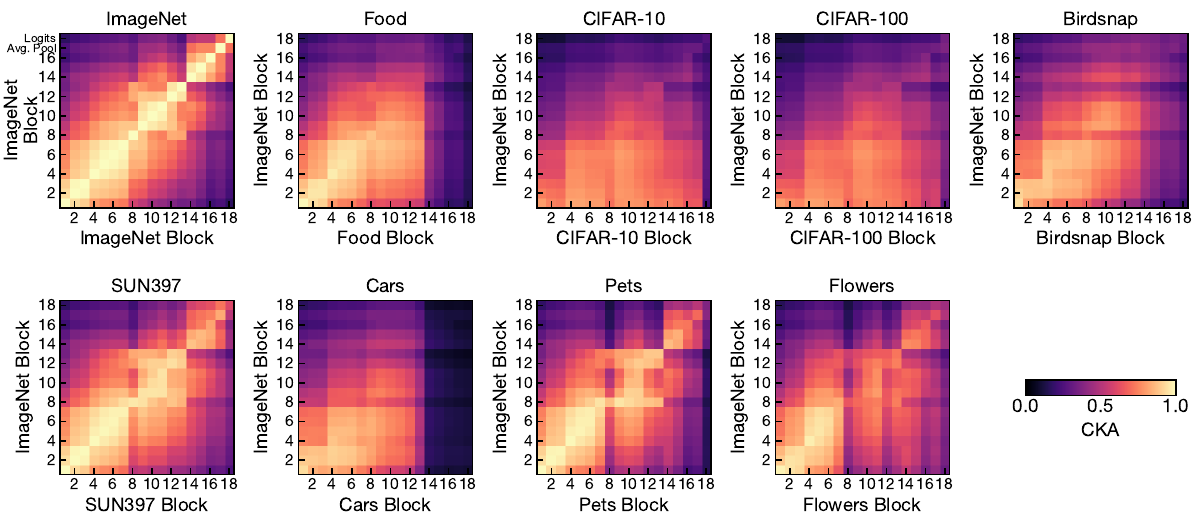}
\end{center}
\vskip -0.5em
\caption{\textbf{Transfer produces large changes in later network layer representations.} We measure CKA between ResNet-50 models trained with softmax cross-entropy before and after transfer, following the same procedure as described in Section~\ref{sec:similarity} of the main text. Consistent with previous work, we find that later layers change more than earlier layers, although interestingly, the extent of the changes differs greatly by dataset.}
\label{fig:cka_transfer}
\vskip -1em
\end{figure}
\vskip -1em

\FloatBarrier
\section{Additional class separation results}
\subsection{Relation of class separation index to variance ratios}
\label{app:relationships}
The class separation index we use is a simple multidimensional generalization of $\eta^2$ in ANOVA or $R^2$ in linear regression with categorical predictors when it is applied to normalized embeddings. Its properties are likely to be familiar to many readers. In this section, for completeness, we derive these properties and provide connections to related work.

The ratio of the average within-class cosine distance to the overall average cosine distance provides a measure of how distributed examples within a class are that is between 0 and 1. We take one minus this quantity to get a closed-form measure of class separation
\begin{align}
    R^2 &= 1 - \frac{\sum_{k=1}^K \sum_{m=1}^{N_k} \sum_{n=1}^{N_k} \left(1 - \mathrm{sim}(\bm{X}^{k}_{m,:}, \bm{X}^k_{n,:})\right)/(KN_k^2)}{\sum_{j=1}^K\sum_{k=1}^K\sum_{m=1}^{N_j} \sum_{n=1}^{N_k} \left(1 - \mathrm{sim}(\bm{X}^{j}_{m,:}, \bm{X}^k_{n,:})\right)/(K^2 N_j N_k)},
\end{align}
where $N_k$ is the number of examples in class $k$, $\bm{X}^k \in \mathbb{R}^{N_k \times P}$ is the matrix of $P$-dimensional embeddings of these examples, and $\mathrm{sim}(\bm{x}, \bm{y}) = \bm{x}^\mathsf{T}\bm{y}/(\|\bm{x}\|\|\bm{y}\|)$ is cosine similarity.

\paragraph{Relation of $R^2$ to ratio of within-class vs. total variance:} $R^2$ is one minus the ratio of the within-class to weighted total variances of $L^2$-normalized embeddings, summed over the feature dimension. To see this, first note that
\begin{align}
    \|\bm{x} / \|\bm{x}\| - \bm{y} / \|\bm{y}\|\|^2 &= \left(\bm{x} / \|\bm{x}\| - \bm{y} / \|\bm{y}\|\right)^\mathsf{T} \left(\bm{x} / \|\bm{x}\| - \bm{y} / \|\bm{y}\|\right)\\
    &= 2 - 2 \mathrm{sim}(\bm{x}, \bm{y}),
\end{align}
so, letting $\tilde{\bm{X}}^k \in \mathbb{R}^{N_k \times P}$ be matrices of $L^2$-normalized embeddings $\tilde{\bm{X}}^k_{m, :} = {\bm{X}}^k_{m, :} / \|\bm{X}^k_{m, :}\|$
\begin{align}
    R^2 &= 1 - \frac{\sum_{k=1}^K \sum_{m=1}^{N_k} \sum_{n=1}^{N_k} \|\tilde{\bm{X}}^k_{m, :} - \tilde{\bm{X}}^k_{n, :}\|^2/(KN_K^2)}{\sum_{j=1}^K\sum_{k=1}^K\sum_{m=1}^{N_j} \sum_{n=1}^{N_k} \|\tilde{\bm{X}}^j_{m, :} - \tilde{\bm{X}}^k_{n, :}\|^2/(K^2 N_j N_k)}.
\end{align}
The variance of a vector is a V-statistic with the kernel $h(x, y) = (x - y)^2 / 2$, i.e.,
\begin{align}
    \mathrm{Var}(\bm{y}) &= \frac{1}{n} \sum_{i=1}^n \left(y_i - \sum_{j=1}^n y_j / n\right)^2
    = \frac{1}{n^2} \sum_{i=1}^n \sum_{j=1}^n (y_i - y_j)^2 / 2,
\end{align}
and thus the sum of the variances of the columns of a matrix $\bm{Y} \in \mathbb{R}^{N \times P}$ is
\begin{align}
    \sum_{p=1}^{P} \mathrm{Var}(\bm{Y}_{:,p}) &= \frac{1}{n^2} \sum_{p=1}^P \sum_{m=1}^N \sum_{n=1}^N (Y_{m, p} - Y_{n, p})^2 / 2 = \frac{1}{n^2} \sum_{m=1}^N \sum_{n=1}^N \|\bm{Y}_{m, :} - \bm{Y}_{n, :}\|^2 / 2.
    \label{eq:variance_formula}
\end{align}
If all $N_k$ are equal\footnote{This equivalence also holds for unequal $N_k$ if the variance is replaced by the inverse-frequency-weighted variance.}, we can use \eqref{eq:variance_formula} to write $R^2$ in terms of the ratio of the average within-class variance to the total variance of the normalized embeddings, where each variance is summed over the embedding dimensions:
\begin{align}
    \sigma^2_\text{within} = \sum_{p=1}^{P}\sum_{k=1}^K  \mathrm{Var}(\tilde{\bm{X}}_{:,p}^k)/K &&
    \sigma^2_\text{total} = \sum_{p=1}^{P}\mathrm{Var}(\tilde{\bm{X}}_{:,p}^\text{all}) && R^2 = 1 - \frac{\sigma^2_\text{within}}{\sigma^2_\text{total}},
\end{align}
where $\tilde{\bm{X}}^\text{all} \in \mathbb{R}^{kN \times P}$ is the matrix of all examples.

\paragraph{Relation of $R^2$ to ratio of between-class vs. total variance:} Letting $\bm{M} \in {K \times P}$ be the matrix of mean normalized embeddings of each class $\bm{M}_{k,:} = \frac{1}{N_k} \sum_{m=1}^{N_k} \tilde{\bm{X}}^{k}_{m,:}$, the law of total variance states that the variance of each dimension is the sum of the within-class and between-class variances:
\begin{align}
    \label{eq:total_var}
    \mathrm{Var}(\tilde{\bm{X}}_{:,p}^\text{all}) &= \sum_{k=1}^K \mathrm{Var}(\tilde{\bm{X}}_{:,p}^k)/K + \mathrm{Var}{(\bm{M}_{:, p})}.
\end{align}
Thus, if we let $\sigma^2_\text{between} = \sum_{p=1}^{P}\mathrm{Var}(\bm{M}_{:, p})$, the variance of the class means summed across dimensions, \eqref{eq:total_var} implies that $\sigma^2_\text{total} = \sigma^2_\text{within} + \sigma^2_\text{between}$. Thus, we have
\begin{align}
    R^2 = \sigma^2_\text{between}/\sigma^2_\text{total}.
\end{align}

\paragraph{Relation of $R^2$ with other variance ratios:} Other work has used the alternative variance ratios $\sigma^2_\text{within}/\sigma^2_\text{between}$~\citep{goldblum2020unraveling}, $\sigma^2_\text{between}/\sigma^2_\text{within}$~\citep{liu2020negative}, or $(\sigma^2_\text{between} - \sigma^2_\text{within})/(\sigma^2_\text{between} + \sigma^2_\text{within})$~\citep{mehrer2020individual} to measure class separation. These ratios are monotonic functions of $R^2$ and can be computed directly from the numbers we provide:
\begin{align}
    \frac{\sigma^2_\text{within}}{\sigma^2_\text{between}} = \frac{1}{R^2} - 1, && \frac{\sigma^2_\text{between}}{\sigma^2_\text{within}} = \frac{R^2}{1 - R^2}, &&
    \frac{\sigma^2_\text{between} - \sigma^2_\text{within}}{\sigma^2_\text{between} + \sigma^2_\text{within}} = 2R^2 - 1.
\end{align}

\FloatBarrier
\subsection{Other class separation indexes and measurements}
\label{app:other_ratios}
\begin{table}[h!]
\caption{\textbf{Comparison of class separation under different distance indexes.} Cosine (mean-subtracted) subtracts the mean of the activations before computing the cosine distance. All results reported for ResNet-50 on the ImageNet training set.}
  \aftertablecaption
\centering
\footnotesize
\begin{tabular}{lrrr}
    \toprule
    Loss/regularizer & \multicolumn{1}{c}{Cosine} & \multicolumn{1}{p{2cm}}{\centering Cosine (mean-subtracted)} & \multicolumn{1}{c}{Euclidean distance} \\
    \midrule
    Softmax & $0.3494 \pm 0.0002$ & $0.3472 \pm 0.0002$ & $0.3366 \pm 0.0002$   \\
    Squared error  & $0.8452 \pm 0.0002$ & $0.8450 \pm 0.0002$ & $0.8421 \pm 0.0007$\\
    Dropout & $0.4606 \pm 0.0003$  & $0.4559 \pm 0.0002$ & $0.4524 \pm 0.0003$ \\
    Label smoothing & $0.4197 \pm 0.0003$ & $0.4124 \pm 0.0004$ & $0.3662 \pm 0.0005$ \\
    Extra final layer $L^2$ & $0.5718 \pm 0.0006$ & $0.5629 \pm 0.0005$ & $0.5561 \pm 0.0005$ \\
    Logit penalty & $0.6012 \pm 0.0004$ & $0.5950 \pm 0.0004$ & $0.5672 \pm 0.0004$ \\
    Logit normalization & $0.5167 \pm 0.0002$ & $0.5157 \pm 0.0002$ & $0.5326 \pm 0.0002$ \\
    Cosine softmax  & $0.6406 \pm 0.0003$ & $0.6389 \pm 0.0003$ & $0.6406 \pm 0.0003$\\
    Sigmoid & $0.4267 \pm 0.0003$ & $0.4315 \pm 0.0003$ & $0.4272 \pm 0.0003$ \\
    \bottomrule
\end{tabular}
\label{tab:r2_alternative}
\end{table}

\begin{table}[h]
\caption{\textbf{Simplex ETF measurements.} \citet{papyan2020prevalence} measure various quantities to demonstrate that the representations of neural networks converge to the simplex equiangular tight frame as the training error goes to 0. Collapse to the equiangular tight frame implies both that class separation is maximal (i.e., $R^2 \to 1$) and that class means are maximally distributed. Here we compute the same quantities for networks trained with different losses and optimal early stopping. $\tilde \mu_c$ indicates the mean embedding of the $c$\textsuperscript{th} of $C = 1000$ total classes after subtracting the global mean across all classes, and $\bm{w}$ indicates the classifier weights corresponding to the $c$\textsuperscript{th} class. $\bm{W}$ is the matrix of all classifier weights, whereas $\bm{M}$ is the matrix of all global-mean-subtracted class mean embeddings. $\bm{\Sigma}_B$ is the covariance matrix of the class means and $\bm{\Sigma}_W$ is the within-class covariance matrix.  $\mathrm{Tr}(\bm{\Sigma}_W \bm{\Sigma}_B^\dagger)/C$, where $\bm{\Sigma}_B^\dagger$ is the Moore-Penrose pseudoinverse of $\bm{\Sigma}_B$, measures the collapse of within-class variability relative to between-class variability. This quantity reveals the greatest difference between softmax and other losses, and it is also the most related to class separation; for isotropic covariance matrices, $\mathrm{Tr}(\bm{\Sigma}_W \bm{\Sigma}_B^\dagger)/C = P/C(1/R^2 - 1)$ where $P$ is the number of penultimate layer features. $\mathrm{Std}_{c,c'}(\mathrm{sim}(\tilde{\bm{\mu}}_c, \tilde{\bm{\mu}}_{c'}))$ and $\mathrm{Avg}_{c,c'}|\mathrm{sim}(\tilde{\bm{\mu}}_c, \tilde{\bm{\mu}}_{c'}) + \frac{1}{C-1}|$, which measure the discrepancy between the class mean directions and those that would result in the maximal separation between the class means, also differentiate softmax from other losses. Other quantities do not.}
\label{tab:simplex_etf}
  \aftertablecaption
    \scriptsize
\hspace{-1cm}
\begin{tabular}{@{}l@{\hspace{.15cm}}llll@{}}
\toprule
{} & $\frac{\mathrm{Std}_c(\|\tilde{\bm{\mu}}_c\|)}{\mathrm{Avg}_c(\|\tilde{\bm{\mu}}_c\|_2)}$ & $\frac{\mathrm{Std}_c(\|\bm{w}_c\|)}{\mathrm{Avg}_c(\|\bm{w}_c\|_2)}$ & $\mathrm{Std}_{c,c'}(\mathrm{sim}(\tilde{\bm{\mu}}_c, \tilde{\bm{\mu}}_{c'}))$ & $\mathrm{Std}_{c,c'}(\mathrm{sim}(\bm{w}_c, \bm{w}_{c'}))$ \\
\midrule
Softmax             &                                 0.127 $\pm$ 0.0007 &                                 0.100 $\pm$ 0.0005 &                        \textbf{0.119} $\pm$ 0.0003 &                                 0.034 $\pm$ 0.0000 \\
Label Smoothing     &                                 0.102 $\pm$ 0.0054 &                                 0.126 $\pm$ 0.0004 &                                 0.096 $\pm$ 0.0010 &                                 0.024 $\pm$ 0.0001 \\
Sigmoid             &                                 0.171 $\pm$ 0.0009 &                                 0.102 $\pm$ 0.0003 &                                 0.080 $\pm$ 0.0005 &                                 0.031 $\pm$ 0.0001 \\
More $L^2$ &                                 0.131 $\pm$ 0.0005 &                                 0.069 $\pm$ 0.0006 &                                 0.086 $\pm$ 0.0003 &                                 0.057 $\pm$ 0.0001 \\
Dropout             &                                 0.134 $\pm$ 0.0012 &                                 0.048 $\pm$ 0.0003 &                                 0.085 $\pm$ 0.0003 &                                 0.061 $\pm$ 0.0001 \\
Logit Penalty       &                                 0.105 $\pm$ 0.0019 &                                 0.092 $\pm$ 0.0002 &                                 0.057 $\pm$ 0.0003 &                                 0.016 $\pm$ 0.0000 \\
Logit Norm &                                 0.300 $\pm$ 0.0014 &                        \textbf{0.135} $\pm$ 0.0007 &                                 0.060 $\pm$ 0.0002 &                                 0.020 $\pm$ 0.0000 \\
Cosine Softmax      &                \textcolor{red}{0.100} $\pm$ 0.0004 &                \textcolor{red}{0.000} $\pm$ 0.0000 &                                 0.045 $\pm$ 0.0002 &                        \textbf{0.063} $\pm$ 0.0001 \\
Squared Error       &                        \textbf{0.448} $\pm$ 0.0050 &                                 0.078 $\pm$ 0.0010 &                \textcolor{red}{0.011} $\pm$ 0.0001 &                \textcolor{red}{0.005} $\pm$ 0.0001 \\
\toprule
{} & $\mathrm{Avg}_{c,c'}|\mathrm{sim}(\tilde{\bm{\mu}}_c, \tilde{\bm{\mu}}_{c'}) + \frac{1}{C-1}|$ & $\mathrm{Avg}_{c,c'}|\mathrm{sim}(\bm{w}_c, \bm{w}_{c'}) + \frac{1}{C-1}|$ & $\|\bm{W}/\|\bm{W}\|_\mathsf{F} - \bm{M}/\|\bm{M}\|_\mathsf{F}\|_\mathsf{F}^2$ & $\mathrm{Tr}(\bm{\Sigma}_W \bm{\Sigma}_B^\dagger)/C$ \\
\midrule
Softmax             &                        \textbf{0.084} $\pm$ 0.0003 &                                 0.025 $\pm$ 0.0000 &                                 0.918 $\pm$ 0.0015 &                      \textbf{21.158} $\pm$ 0.1283 \\
Label Smoothing     &                                 0.068 $\pm$ 0.0008 &                                 0.018 $\pm$ 0.0001 &                                 0.954 $\pm$ 0.0052 &                                9.001 $\pm$ 0.0859 \\
Sigmoid             &                                 0.057 $\pm$ 0.0005 &                        \textbf{0.056} $\pm$ 0.0004 &                        \textbf{0.960} $\pm$ 0.0013 &                                6.957 $\pm$ 0.0371 \\
More $L^2$ &                                 0.059 $\pm$ 0.0003 &                                 0.042 $\pm$ 0.0001 &                                 0.328 $\pm$ 0.0009 &                                6.104 $\pm$ 0.1007 \\
Dropout             &                                 0.057 $\pm$ 0.0003 &                                 0.045 $\pm$ 0.0001 &                                 0.521 $\pm$ 0.0010 &                                9.701 $\pm$ 0.0494 \\
Logit Penalty       &                                 0.034 $\pm$ 0.0003 &                                 0.010 $\pm$ 0.0000 &                                 0.481 $\pm$ 0.0020 &                                2.722 $\pm$ 0.2005 \\
Logit Norm &                                 0.034 $\pm$ 0.0002 &                                 0.013 $\pm$ 0.0000 &                                 0.759 $\pm$ 0.0016 &                                2.943 $\pm$ 0.1150 \\
Cosine Softmax      &                                 0.023 $\pm$ 0.0002 &                                 0.045 $\pm$ 0.0001 &                \textcolor{red}{0.302} $\pm$ 0.0004 &                                1.305 $\pm$ 0.3153 \\
Squared Error       &                \textcolor{red}{0.002} $\pm$ 0.0000 &                \textcolor{red}{0.002} $\pm$ 0.0000 &                                 0.647 $\pm$ 0.0043 &               \textcolor{red}{0.334} $\pm$ 0.0260 \\

\bottomrule
\end{tabular}

\end{table}

\begin{figure}[h!]
\begin{center}
      \includegraphics[width=\textwidth]{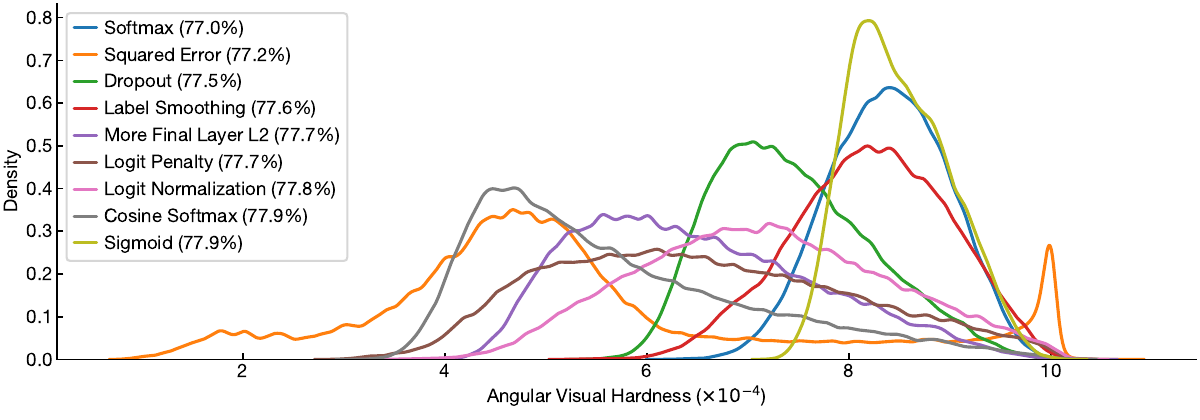}
\end{center}
\vskip -1.0em
\caption{\textbf{Angular visual hardness of different loss functions.} Kernel density estimate of the angular visual hardness~\citep{chen2019angular} scores of the 50,000 examples in the ImageNet validation set, computed with a Gaussian kernel of bandwidth $5 \times 10^{-6}$, for ResNet-50 networks trained with different losses. Legend shows ImageNet top-1 accuracy for each loss function in parentheses. Although alternative loss functions generally reduce angular visual hardness vs. softmax loss, sigmoid loss does not, yet it is tied for the highest accuracy of any loss function.}
\label{fig:avh}
\end{figure}

\begin{figure}[h!]
\begin{center}
      \includegraphics[width=0.95\textwidth]{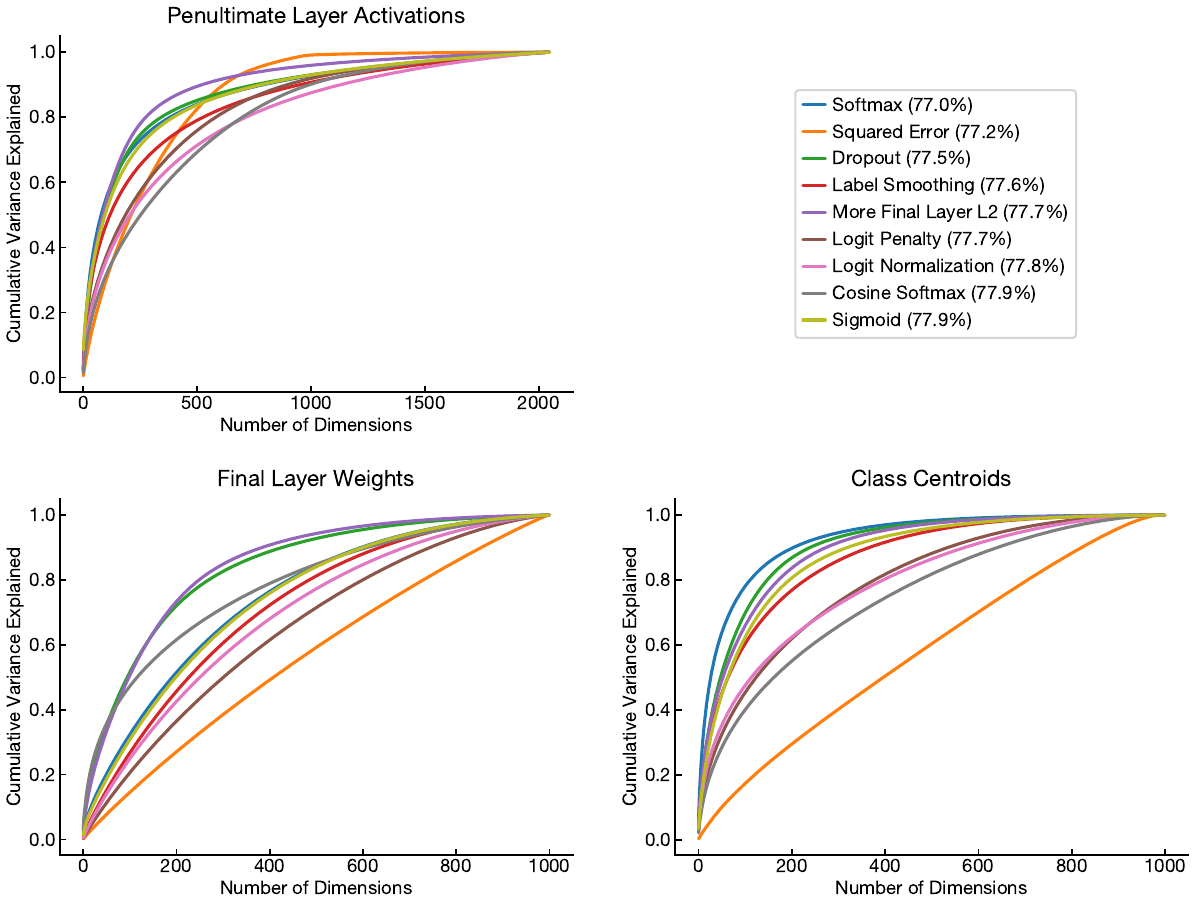}
\end{center}
\vskip -0.8em
\caption{\textbf{Singular value spectra of activations and weights learned by different losses.} Singular value spectra computed for penultimate layer activations, final layer weights, and class centroids of ResNet-50 models on the ImageNet training set. Penultimate layer activations and final layer weights fail to differentiate sigmoid cross-entropy from softmax cross-entropy. By contrast, the singular value spectrum of the class centroids clearly distinguishes these losses.}
\label{fig:eigenspectra}
\vskip -1em
\end{figure}

\begin{figure}[h!]
\begin{center}
      \includegraphics[width=0.95\textwidth]{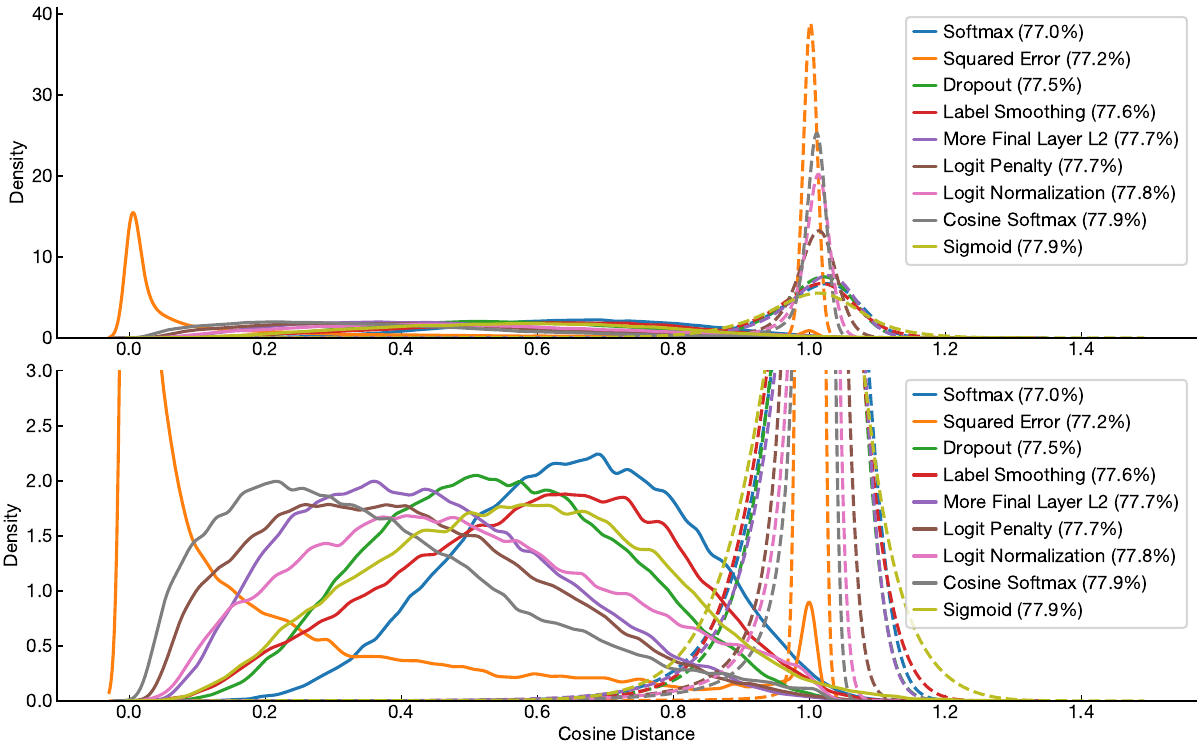}
\end{center}
\vskip -0.8em
\caption{\textbf{The distribution of cosine distance between examples.} Kernel density estimate of the cosine distance between examples of the same class (solid lines) and of different classes (dashed lines), for penultimate layer embeddings of 10,000 training set examples from ResNet-50 on ImageNet. Top and bottom plots show the same data with different y scales.}
\label{fig:cosine_dist}
\end{figure}

\FloatBarrier
\subsection{Augmentation can improve accuracy without increasing class separation}
In Section~\ref{app:combined_losses}, we show that improved loss functions and AutoAugment are additive, whereas combinations of improved loss functions or regularizers lead to no significant accuracy improvements. In Table~\ref{tab:autoaugment_class_separation} below, we show that AutoAugment also does not increase class separation. These results confirm that data augmentation and modifications to networks' final layers exert their effects via different (and complementary) mechanisms.
\label{app:autoaugment_class_separation}
\begin{table}[h!]
    \centering
  \caption{\textbf{AutoAugment increases ImageNet top-1 accuracy without increasing class separation.} Top-1 accuracy is computed on the ImageNet validation set; class separation is computed on the ImageNet training set. Results are averaged over 3 (with AutoAugment) or 8 (standard augmentation) models.}
  \aftertablecaption
  \begin{tabular}{lrrrrr}
    \toprule
    & \multicolumn{2}{c}{Standard augmentation} & \multicolumn{2}{c}{AutoAugment}\\
    \cmidrule(lr){2-3} \cmidrule(lr){4-5}
    Loss & \multicolumn{1}{p{2cm}}{\centering ImageNet top-1} & \multicolumn{1}{p{2cm}}{\centering Class sep. ($R^2$)} & \multicolumn{1}{p{2cm}}{\centering ImageNet top-1} & \multicolumn{1}{p{2cm}}{\centering Class sep. ($R^2$)}\\
    \midrule
    Softmax & 77.0 $\pm$ 0.06 & 0.349 $\pm$ 0.0002 & 77.7 $\pm$ 0.05 & 0.353 $\pm$ 0.0002\\
    Sigmoid & 77.9 $\pm$ 0.05 & 0.427 $\pm$ 0.0003 & 78.5 $\pm$ 0.04 & 0.432 $\pm$ 0.0001\\
    Logit penalty & 77.7 $\pm$ 0.04 & 0.601 $\pm$ 0.0004 & 78.3 $\pm$ 0.05 & 0.595 $\pm$ 0.0003\\
    Cosine softmax & 77.9 $\pm$ 0.02 & 0.641 $\pm$ 0.0003 & 78.3 $\pm$ 0.05 & 0.632 $\pm$ 0.0001\\
    \bottomrule
  \end{tabular}
  \label{tab:autoaugment_class_separation}
\end{table}

\FloatBarrier
\subsection{Training dynamics of class separation}
\label{app:training_dynamics}
\begin{figure}[h!]
\begin{center}
      \includegraphics[width=\textwidth]{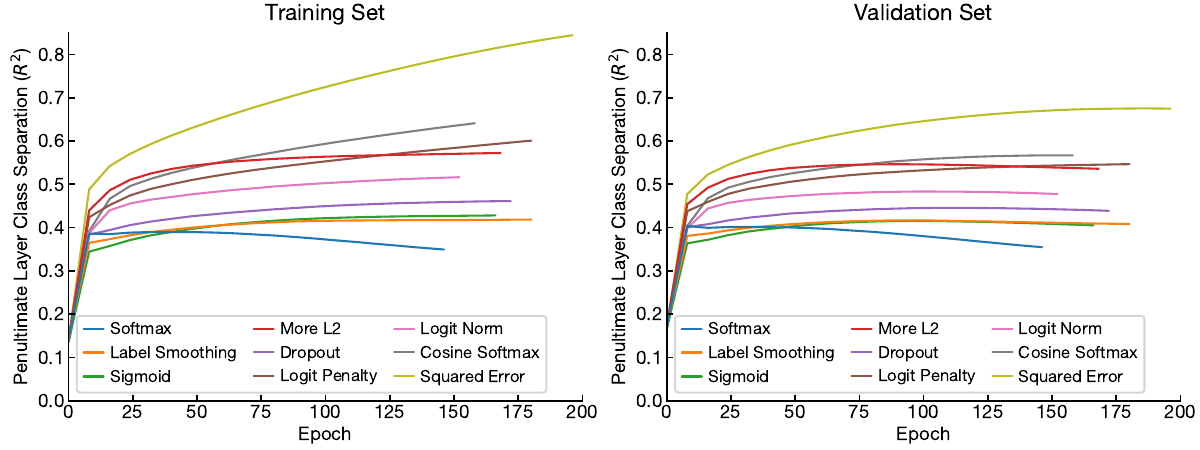}
\end{center}
\vskip -1em
\caption{\textbf{Differences in training dynamics of class separation across loss functions.} Plots show evolution of class separation over training on the ImageNet training set (left) and validation set (right), for a single ResNet-50 model of each loss function type evaluated every 8 epochs. Curves terminate at the epoch that provided the highest holdout set accuracy.}
\label{fig:class_separation_over_training}
\end{figure}

As we discuss in detail in Section~\ref{sec:separation} of the main text, different loss functions lead to different values of class separation. However, we train models with different loss functions for different numbers of epochs, reported in Table~\ref{tab:imagenet_hyperparameters}. In this section, we confirm that differences in the number of training epochs alone do not explain differences in observed class separation among losses. Instead, as shown in Figure~\ref{fig:class_separation_over_training}, differences among losses are established early in training and the relative ordering changes little. We observe that, for the softmax cross-entropy model, class separation peaks at epoch 32 and then falls, whereas all models trained with different objectives achieve maximum class separation on the training set at the last checkpoint. On the validation set, for most losses, class separation peaks before optimal accuracy is reached.

\FloatBarrier
\section{Class overlap between ImageNet and transfer datasets}
\label{app:class_level_overlap}
In this section, we investigate overlap in the classes contained in the ImageNet ILSVRC 2012 dataset and those contained in the downstream datasets investigated in this work. We previously reported the overlap in the \textit{images} contained in these datasets and those contained in ImageNet in Appendix H of~\citet{kornblith2019better}.

We first measure the number of classes in each dataset where the class names correspond semantically to classes in ImageNet. Due to differences in the granularity of different datasets, semantic class overlap is somewhat ambiguous. For example, CIFAR-10 contains a single ``dog'' class that corresponds to 90 dog breeds contained in ImageNet, but ImageNet contains only a single ``hummingbird'' class whereas Birdsnap contains 9 different species. We consider classes as ``overlapping'' when the name of the downstream either directly or nearly corresponds to an ImageNet class, or is a superclass of ImageNet classes.

In addition to being ambiguous, semantic class overlap does not consider shift in class-conditional distributions. Simply because classes in two datasets refer to the same kinds of real-world objects does not mean that the images those classes contain are similar. For example, 61 of the 100 classes in CIFAR-100 are superclasses of ImageNet classes, but because CIFAR images are much lower resolution, a classifier trained on ImageNet does not perform well at classifying them.

To develop a measure of class overlap that takes distribution shift into consideration, we map each ImageNet class to a class in the downstream dataset and use this mapping in combination with the original 1000-way ImageNet-trained vanilla softmax cross-entropy network to measure classification accuracy. Finding the optimal class mapping is an instance of the minimum-cost flow problem, but can also be solved somewhat less efficiently as a variant of the assignment problem. For each downstream task, we apply an ImageNet classifier to the task's training set and compute the matching matrix. The cost matrix for the assignment problem is the negative matching matrix. To allow multiple ImageNet classes to be assigned to the same downstream task class, we replicate all classes in the downstream task $k$ times, where we select $k$ so that $k+1$ replications provides no improvement in accuracy, then use \verb!scipy.optimize.linear_sum_assignment! to find the mapping. We call the accuracy of the resulting mapping the ``assignment accuracy.''

Results are shown in Table~\ref{tab:class_overlap}. Semantic class overlap is generally low, but CIFAR-10, CIFAR-100, and Pets all have non-trivial semantic class overlap. For most datasets, the assignment accuracy is less than half the linear transfer accuracy; the only exceptions are CIFAR-10 and Pets, where the drop is smaller. Pets has comparable linear transfer accuracy to CIFAR-10 but higher assignment accuracy, and thus arguably has the greatest class overlap with ImageNet.

\begin{table}[h!]
    \centering
  \caption{\textbf{Class overlap between ImageNet and transfer datasets.}}
  \aftertablecaption
  \begin{tabular}{lrrrrr}
    \toprule
    Dataset & \multicolumn{1}{p{1.5cm}}{\centering Number of classes} & \multicolumn{1}{p{1.7cm}}{\centering Semantic class overlap} & \multicolumn{1}{p{1.5cm}}{\centering Assignment accuracy} & \multicolumn{1}{p{1.9cm}}{\centering Linear transfer accuracy}\\
    \midrule
    Food & 101 & 7 & 15.4 & 74.6\\
    CIFAR-10 & 10 & 9 & 65.1 & 92.4\\
    CIFAR-100 & 100 & 61 & 34.6 & 76.9\\
    Birdsnap & 500 & 11 & 7.0 & 55.4\\
    SUN397 & 397 & 39 & 20.2 & 62.0\\
    Cars & 196 & 0 & 4.0 & 60.3\\
    Pets & 37 & 25 & 71.2 & 92.0\\
    Flowers & 102 & 1 & 7.6 & 94.0\\
    \bottomrule
  \end{tabular}
  \label{tab:class_overlap}
\end{table}

\end{bibunit}

\end{document}